\documentclass{article}

 \usepackage[preprint]{neurips_2026}

\usepackage[utf8]{inputenc} 
\usepackage[T1]{fontenc}    
\usepackage{hyperref}       
\usepackage{url}            
\usepackage{tabularx}
\usepackage{booktabs}       
\usepackage{amsfonts}       
\usepackage{nicefrac}       
\usepackage{microtype}      
\usepackage{xcolor}         

\usepackage{color, soul}
\usepackage{colortbl}
\usepackage{xfp}
\usepackage{microtype}
\usepackage{graphicx}
\usepackage{subcaption}
\usepackage{amsmath}
\usepackage{amssymb}
\usepackage{mathtools, commath}
\usepackage{tcolorbox}
\tcbuselibrary{skins}
\usepackage{amsthm}
\usepackage{bbm}
\usepackage{makecell}
\usepackage{multirow}
\usepackage{mathtools}
\usepackage{algorithm}
\usepackage{algorithmic}
\usepackage{wrapfig,lipsum}
\usepackage{comment}
\usepackage{adjustbox}
\usepackage{threeparttable}
\usepackage{array}

\usepackage{pifont}
\newcommand{\cmark}{\ding{51}}%
\newcommand{\xmark}{\ding{55}}%
\usepackage[shortlabels]{enumitem}
\usepackage[capitalize,noabbrev]{cleveref}
\newcommand{\ms}[2]{{#1}\scriptsize{$\pm$#2}}
\newcommand{\msone}[2]{\textbf{{#1}}\scriptsize{$\pm$#2}}
\newcommand{\mstwo}[2]{\underline{{#1}\scriptsize{$\pm$#2}}}

\usepackage{enumitem}

\usepackage{CJKutf8}

\theoremstyle{plain}

\theoremstyle{definition}

\theoremstyle{remark}

\usepackage[normalem]{ulem}
\useunder{\uline}{\ul}{}

\usepackage{pifont}
\newcommand{\barwidth}{7} %
\newcommand{\barheight}{7.5pt} %
\newcommand{\percentscale}{100} %
\definecolor{myred}{RGB}{213,122,122} %
\def\pcb#1{%
   {\color{myred}\rule{\fpeval{#1/\percentscale*\barwidth} cm}{\barheight}} #1
}

\title{TTGBench: Benchmarking Topological Evolution and Semantic Drift in Text-attributed Temporal Graphs}

\author{Longfei Ma$^1$, Zemin Liu$^{1}$\thanks{Corresponding author} ,  Fei Wu$^1$\footnotemark[1]\\
$^1$Zhejiang University\\
\texttt{\{longfeima, liu.zemin, wufei\}@zju.edu.cn,}\\
}

\begin{document}

\maketitle

\begin{abstract}
Temporal graph learning models the evolution of dynamic systems, where both structural interactions and semantic states change over time. However, existing benchmarks primarily emphasize structural evolution via temporal link prediction (TLP), while support for semantic evolution remains limited. Although temporal node classification (TNC) is sometimes included, it is typically restricted to simplistic binary settings that fail to capture realistic semantic drift. Moreover, commonly used datasets exhibit high link repetition, leading to inflated performance estimates and obscuring true model capability.
To address these limitations, we introduce \textbf{TTGBench}, a new benchmark that jointly evaluates structural and semantic evolution. TTGBench comprises six real-world, text-rich datasets characterized by \emph{Dual Volatility}, enabling rigorous and fair evaluation of existing models. Notably, it is the first benchmark to support both multi-class and multi-label TNC, filling a critical gap in evaluating temporal semantic drift.
We conduct a comprehensive evaluation of 17 state-of-the-art methods across Temporal Graph Neural Networks (TGNNs) and Large Language Model (LLM)-based paradigms. The results reveal a clear \emph{capability divide} between the two paradigms: TGNN-based methods excel at structural prediction but fail at semantic tracking, whereas LLM-based predictors show the opposite trend. Through in-depth analysis, we uncover their fundamental limitations and provide insights for developing more comprehensive temporal graph models. 

\end{abstract}


\section{Introduction}
\label{sec:intro}

Temporal graphs provide a powerful framework for modeling dynamic systems such as social networks\cite{zhang2023location,dileo2024temporal,mitra2025analyzing}, e-commerce platforms\cite{ding2019user,zhao2024dynllm}, and financial transaction networks\cite{zhou2023graph,trinh2024dynamic}. At their core, these systems evolve along two fundamental dimensions: \emph{structural evolution} and \emph{semantic drift}. Structural evolution describes how nodes form and dissolve connections over time, typically modeled as \emph{temporal link prediction} (TLP). In parallel, semantic drift captures how node labels or roles change over time—such as shifts in user interests—and is commonly formulated as \emph{temporal node classification} (TNC). A comprehensive understanding of both dimensions is essential for uncovering the underlying dynamics of temporal graphs. Therefore, a benchmark that jointly evaluates TLP and TNC is crucial for assessing whether models can capture the full spectrum of temporal evolution.

Existing benchmarks have significantly advanced temporal graph learning by curating diverse datasets (e.g., DGB~\cite{poursafaei2022towards}, TGB~\cite{huang2023temporal}), introducing text-attributed graphs (DTGB~\cite{zhang2024dtgb}), and establishing unified evaluation pipelines (DyGLib~\cite{yu2023towards}). Despite these contributions, two fundamental limitations remain. First, most benchmarks focus almost exclusively on structural evolution through TLP, while largely overlooking semantic drift. Although DyGLib supports TNC, it is restricted to simple binary classification tasks (e.g., whether a user is banned), which fail to capture meaningful and continuous semantic drift. This limitation largely stems from the scarcity of datasets with rich, dynamically evolving semantic labels, as noted in prior work~\cite{JMLR:v21:19-447,longa2023graph}. Second, existing datasets often contain a high proportion of repeated edges, leading to overestimated model performance. In fact, many methods achieve near-saturated results (e.g., over 98\% on widely used benchmarks~\cite{yu2023towards,zhang2024dtgb}), which obscures true model capability and fails to reflect realistic, high-novelty environments—a concern also highlighted by recent study~\cite{yi2025tgbseq}.

\begin{wrapfigure}{r}{0.5\textwidth}
    \centering
    \includegraphics[width=\linewidth]{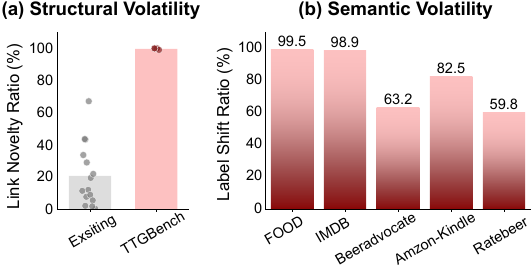}
    \caption{Illustration of dual volatility in TTGBench: structural and semantic volatility.
(a) The proposed datasets exhibit high novelty, where previously unseen edges continuously emerge over time, leading to pronounced structural volatility.
(b) A substantial proportion of nodes change their labels over time, highlighting the presence of semantic volatility in temporal graphs.}
    \label{fig:teaser}
\end{wrapfigure}

To address these limitations, we introduce the Text-attributed Temporal Graph Benchmark (TTGBench), a new benchmark that jointly evaluates structural and semantic evolution. TTGBench comprises six real-world datasets spanning diverse domains and is characterized by a novel property we term \emph{Dual Volatility} (\cref{fig:teaser}): (i) \textit{Structural Volatility}, defined by high link novelty, limited repetition, and highly dynamic interaction patterns; and (ii) \textit{Semantic Volatility}, where node labels evolve frequently with complex, non-trivial dynamics. These properties make TTGBench substantially more challenging than existing benchmarks. Under high structural volatility, memorization-based methods such as EdgeBank~\cite{poursafaei2022towards} fail due to the scarcity of repeated interactions, as confirmed in \cref{sec:exp}. Beyond global novelty, our datasets exhibit intricate structural evolution patterns (\cref{subsec:stru_vola}), requiring models to capture fine-grained temporal dependencies. On the semantic side, TTGBench initially supports both multi-class and multi-label TNC, enabling realistic and diverse semantic scenarios with complex label transitions (\cref{subsec:seman_vola}). Furthermore, all datasets include rich textual attributes, providing essential context for modeling both structural and semantic dynamics.

Building on TTGBench, we conduct a comprehensive evaluation of 17 state-of-the-art methods spanning different paradigms, including dominant TGNNs and recent LLM-based approaches. We categorize these methods into two groups based on the predictor: \emph{TGNN-Predictors}, which include both pure TGNNs and LLM-as-Enhancer methods, and \emph{LLM-Predictors}, which directly use LLMs for prediction. Our evaluation reveals a striking \emph{capability divide}: TGNN-Predictors excel at TLP but usually fail on TNC, whereas LLM-Predictors perform strongly on TNC but struggle with TLP. This dichotomy exposes a fundamental limitation—neither paradigm can simultaneously model structural and semantic evolution. Our empirical results further reveal inherent limitations in how LLM-Predictors encode structural information. In addition, we conduct in-depth analyses to diagnose the failure of TGNNs on semantic tracking, identifying three intrinsic architectural bottlenecks that hinder their ability to utilize semantic signals. Finally, we perform a comprehensive efficiency analysis, providing practical insights into the deployment trade-offs of these methods. Collectively, our findings offer a deeper understanding of current approaches and provide guidance for designing future models that unify structural and semantic learning in temporal graphs.


We summarize our contributions as follows:
\begin{itemize}[leftmargin=*, itemsep=0.1pt,topsep=0.1pt]
\item \textbf{A unified benchmark:}
We introduce \textbf{TTGBench}, the first benchmark that jointly evaluates structural evolution and semantic tracking. It comprises six real-world, text-attributed datasets exhibiting \emph{Dual Volatility}, enabling a more rigorous and holistic assessment of existing models.
\item \textbf{Comprehensive empirical evaluation:}
We conduct a large-scale evaluation of 17 methods spanning both \textbf{TGNN-Predictors} and \textbf{LLM-Predictors}. Our results reveal a clear \emph{capability divide}: TGNN-Predictors excel at TLP but struggle with TNC, whereas LLM-Predictors show the opposite trend, exposing a fundamental limitation of current paradigms.
\item \textbf{Diagnostic empirical insights:}
We provide in-depth analyses to uncover the root causes of these limitations, including structural deficiencies in LLM-based predictors and architectural bottlenecks in TGNNs for semantic tracking, along with their performance and efficiency trade-offs. These findings offer actionable insights for developing more effective temporal graph models.
\end{itemize}

\section{Related Work}

\textbf{Temporal Graph Learning.}  
Temporal graph learning has attracted increasing attention due to its strong ability to model dynamic real-world systems~\cite{kazemi2020representation,skarding2021foundations}. Among existing approaches, TGNNs have emerged as the dominant paradigm owing to their expressive power~\cite{longa2023graph}. In particular, continuous-time temporal graphs, which model interactions with arbitrary timestamps, provide greater flexibility and have enabled many state-of-the-art methods~\cite{kumar2019predicting,trivedi2019dyrep,xu2020inductive,cong2023we,yu2023towards}. Despite this progress, most studies focus primarily on temporal link prediction. To obtain a more comprehensive understanding of model capabilities, we additionally evaluate these methods on multi-class and multi-label temporal node classification.

\textbf{Temporal Graph Benchmarks.}  
Existing benchmarks have substantially advanced temporal graph research by providing diverse datasets and standardized evaluation protocols. DGB~\cite{poursafaei2022towards}, DyGLib~\cite{yu2023towards}, TGB~\cite{huang2023temporal}, and DGraph~\cite{huang2022dgraph} offer unified pipelines for training and evaluation, while DTGB~\cite{zhang2024dtgb} introduces text-attributed temporal graph datasets. TGB-Seq~\cite{yi2025tgbseq} further identifies excessive edge repetition in prior benchmarks and proposes datasets with higher novelty. However, these benchmarks mainly emphasize link-level tasks and, at the node level, are largely limited to binary classification, overlooking more realistic multi-class scenarios. In contrast, our benchmark introduces text-rich temporal graph datasets with high link novelty and supports both multi-class and multi-label node classification within a unified framework.

\textbf{LLMs for Temporal Graph Learning.}  
With the rapid advancement of large language models (LLMs), recent work has begun exploring their use in graph learning. LLM-as-Enhancer methods~\cite{roy2025llm,zhang2025unifying} integrate LLMs to improve temporal graph models, while TGTalker~\cite{huang2025are} employs LLMs as predictors via in-context learning. However, most LLM-based graph learning studies focus on static graphs. To extend LLMs to temporal settings, we adapt recent LLM-as-Predictor methods originally designed for static graphs~\cite{chen2024llaga,tang2024graphgpt} to temporal graphs, enabling empirical evaluation of their ability to directly model both structural and semantic dynamics.

\section{Task Formulation}
\label{sec:task}

To comprehensively evaluate model capabilities in capturing both topological and semantic dynamics, TTGBench standardizes two fundamental tasks. We begin by formally defining the data structure.

\textbf{Definition 1: Text-Attributed Temporal Graph.}
A text-attributed temporal graph is defined as $\mathcal{G} = (\mathcal{V}, \mathcal{E}, \mathcal{T}, \mathcal{X}, \mathcal{Y})$, where $\mathcal{V}$ denotes the set of nodes and $\mathcal{E}$ is a chronologically ordered sequence of timestamped interactions. Each interaction is represented as an event $e = (u, v, t, x_e) \in \mathcal{E}$, indicating that nodes $u, v \in \mathcal{V}$ interact at time $t \in \mathcal{T}$, accompanied by an edge-level textual attribute $x_e$. In addition, each node $u$ is associated with a time-dependent semantic label $y_u(t) \in \mathcal{Y}$, reflecting its instantaneous preference.

Based on this formulation, TTGBench defines two continuous-time tasks to evaluate a model’s ability to capture structural evolution and semantic dynamics.

\textbf{Temporal Link Prediction (TLP): Modeling Structural Evolution.}
TLP evaluates a model’s ability to predict future interactions based on historical graph observations. Given the graph up to time $t$, denoted as $\mathcal{G}(\le t)$, the objective is to predict whether an edge $(u, v)$ will occur at a future time $t' > t$, i.e., whether $(u, v, t') \in \mathcal{E}$. Formally:
\begin{equation}
P(e = (u, v, t') \mid \mathcal{G}(\le t)) = \sigma\left( f_\theta^{(\text{link})}(u, v, \mathcal{G}_{\le t}) \right),
\end{equation}
where $\sigma(\cdot)$ is the sigmoid function mapping model outputs to link probabilities.

\textbf{Temporal Node Classification (TNC): Modeling Semantic Evolution.}
In contrast to prior benchmarks that restrict node classification to simplistic, quasi-static binary tasks, TTGBench formulates TNC as a dynamic semantic prediction problem. As nodes interact over time, their semantic labels $y_u(t)$ evolve continuously. Given the historical graph observed up to time $t$, the goal is to predict node u’s instantaneous semantic label $y_u(t')$ at a future time $t' > t$.

Importantly, TTGBench supports both multi-class and multi-label settings, enabling the modeling of complex and realistic semantic dynamics. The task is defined as:
\begin{equation}
P(y_{u}(t') = c \mid u, \mathcal{G}{\le t}) = \text{Softmax}\left( f_\theta^{(\text{node})}(u, \mathcal{G}{\le t}) \right)_c \quad \text{(multi-class)},
\end{equation}
\begin{equation}
P(\mathbf{Y}_{u, c}(t') = 1 \mid u, \mathcal{G}{\le t}) = \sigma\left( f_{\theta, c}^{(\text{node})}(u, \mathcal{G}{\le t}) \right) \quad \text{(multi-label)},
\end{equation}
where $\mathbf{Y}_{u}(t') \in \{0,1\}^{|\mathcal{C}|}$ is the ground-truth label vector, $\mathcal{C}$ denotes the label set, and $c \in \mathcal{C}$.

\section{The Proposed Datasets: Unveiling Dual Volatility}
\label{sec:datasets}

To rigorously evaluate the capabilities of temporal graph learning models in highly dynamic environments, we construct six real-world, text-attributed temporal graph datasets spanning diverse domains, including culinary recipe feedback~\cite{majumder2019generating}, movie reviews~\cite{misra2019imdb}, book reading records~\cite{cai2017spmc,zhao2015improving}, beer rating data~\cite{mcauley2012learning,mcauley2013amateurs}, and online shopping interactions~\cite{hou2024bridging}. Key statistics are summarized in Table~\ref{tab:datasets}, with additional details provided in Appendix~\ref{appx:dataset}.

Our datasets are fundamentally more challenging than existing benchmarks, as they exhibit a property we term \emph{Dual Volatility}: (i) \textit{Structural Volatility}, characterized by continuous and drastic reshaping of network topology, and (ii) \textit{Semantic Volatility}, characterized by pervasive and complex semantic drift. Beyond the intuitive description introduced in~\cref{sec:intro}, we provide a fine-grained, and micro-level analysis of these properties in this section.

\begin{table*}[t]
    \centering
    \caption{Summary statistics of TTGBench datasets. $C$ denotes node class count.}
    \label{tab:datasets}
    \begin{adjustbox}{width=0.9 \textwidth}
    \begin{tabular}{crrrccc}
        \toprule
        Dataset & \# Nodes & \# Edges & \# Steps & Novelty  & \# $C$ & Tasks \\
        \midrule
        FOOD & 33,773 & 401,057 & 6,252 & 1.0 & 9 & TLP \& multi-label TNC\\ 
        IMDB & 32,371 & 310,891 & 7,099 & 1.0 & 8 & TLP \& multi-label TNC\\ 
        Librarything & 51,201 & 785,690 & 2,914 & 1.0 & - & TLP \\ 
        Beeradvocate & 99,361 & 1,586,573 & 1,577,920 & 0.99 & 7 & TLP \& multi-class TNC\\ 
        Ratebeer & 139,538 & 2,924,105 & 4,254 & 1.0 & 8 & TLP \& multi-class TNC\\ 
        Amazon-Kindle & 215,504 & 5,621,343 & 5,484,604 & 1.0 & 8 & TLP \& multi-class TNC\\ 
        \bottomrule
    \end{tabular}
    \end{adjustbox}
\end{table*}

\subsection{Structural Volatility}
\label{subsec:stru_vola}

\begin{figure}[t]
  \centering
  \includegraphics[width=1.\linewidth]{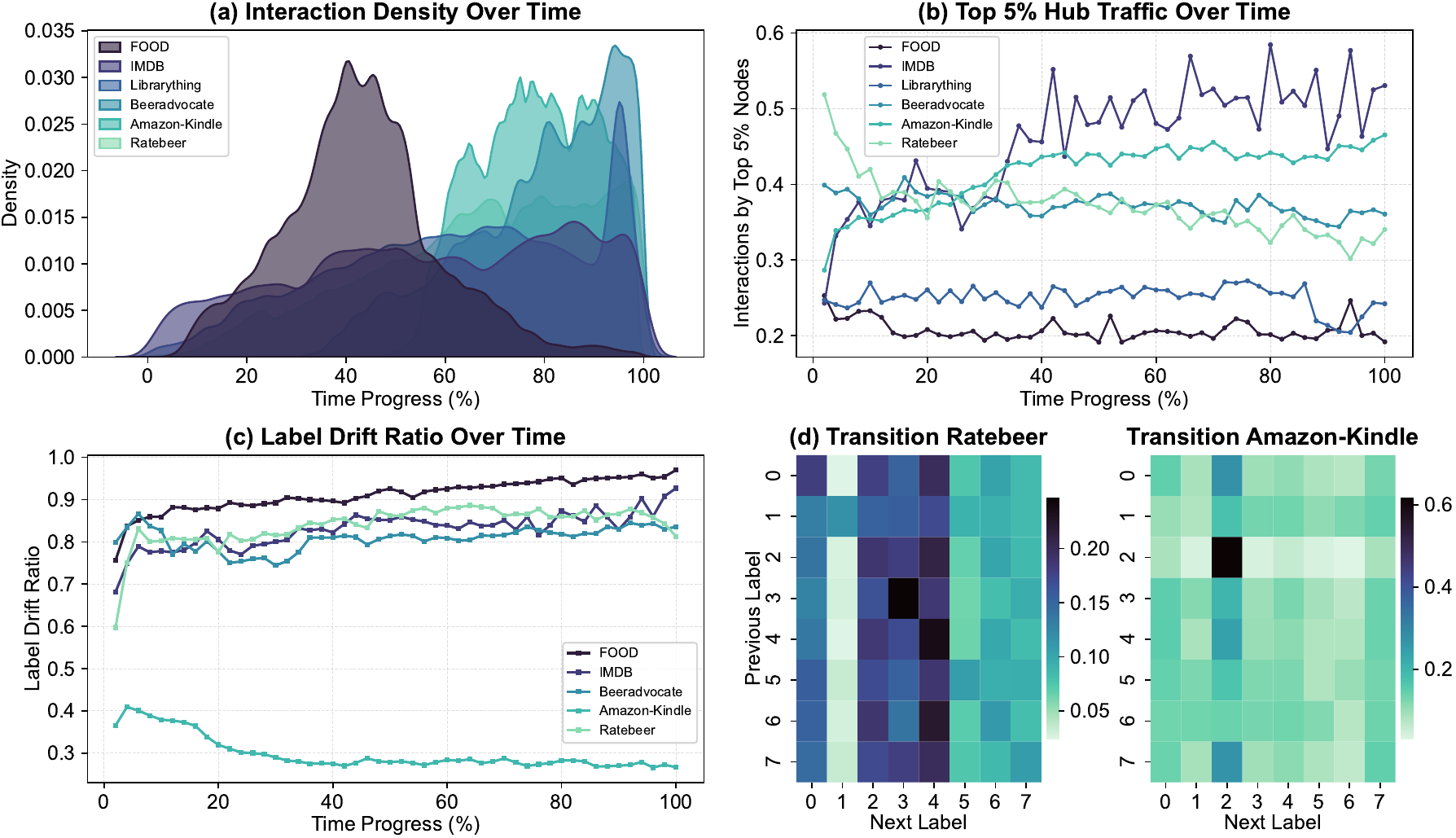}
  \caption{Micro-level visualization of dual volatility across the proposed datasets.
(a) Interaction Density Over Time: Shows the temporal evolution of interaction density, where higher values indicate more frequent and concentrated interactions at a given time.
(b) Top 5\% Hub Traffic Over Time: Illustrates the temporal dynamics of interactions involving the top 5\% most active nodes (hubs).
(c) Label Drift Ratio Over Time: Depicts the proportion of nodes whose labels change over time.
(d) Transition Matrices: Visualizes label transitions on two representative datasets, RateBeer and Amazon-Kindle, where the vertical axis denotes node labels at the previous time step, and the horizontal axis denotes labels at the subsequent time step.}
  \label{fig:dual_volatility}
\end{figure}

We analyze structural dynamics from two complementary perspectives: continuous density variation and traffic concentration instability, as illustrated in~\cref{fig:dual_volatility} (a) and (b).

\textbf{Continuous Density Variation.}
Rather than remaining stable or growing uniformly, interaction volumes exhibit persistent and significant fluctuations over time. This continuous variation indicates that the temporal graphs undergo repeated phases of expansion and contraction. As a result, models must maintain strong temporal adaptability, avoiding overfitting to dense intervals while remaining robust during sparse periods.

\textbf{Traffic Concentration Instability.}
\cref{fig:dual_volatility} (b) shows that the top 5\% most active nodes (hubs) exhibit highly irregular and volatile activity patterns. The graphs continuously shift between centralized regimes—where a small number of nodes dominate interactions—and decentralized regimes with more evenly distributed activity. In other words, the “centers of gravity” of the network are constantly shifting. This instability causes structural patterns learned in one time window to quickly become obsolete, requiring models to adapt to highly non-stationary degree distributions.

\subsection{Semantic Volatility}
\label{subsec:seman_vola}

A key distinction between temporal and static graphs lies in the dynamic nature of node semantics: node labels can evolve over time. We next examine the semantic dynamics of our datasets in detail.

\textbf{High-Frequency Semantic Drifts.}
\cref{fig:dual_volatility} (c) shows that frequent semantic drifts are a common phenomenon across datasets, although their intensity varies by domain. In datasets such as FOOD and IMDB, the drift ratio remains consistently high (often above 0.8), indicating rapid and continuous changes in user preferences. In contrast, Amazon-Kindle exhibits a more moderate yet still substantial drift ratio (around 0.3), reflecting longer periods of semantic consistency. This variation aligns with domain-specific behaviors: while dining and movie consumption are fast-paced, reading typically involves sustained engagement, leading to slower semantic transitions. Importantly, this spectrum—from highly dynamic to moderately stable—demonstrates the diversity of semantic evolution patterns in our datasets, enabling a comprehensive evaluation of models under varying degrees of semantic volatility.

\textbf{Non-Trivial Evolutionary Trajectories.}
\cref{fig:dual_volatility} (d) illustrates label transitions for the same node across consecutive time steps. The resulting heatmaps exhibit asymmetric, high-density clusters rather than uniform or trivially dominant patterns. This indicates that semantic evolution involves abrupt and heterogeneous transitions across distinct classes. Consequently, future states cannot be reliably inferred by simply copying the most recent label or applying a static global transition pattern, highlighting the intrinsic complexity of semantic drift in temporal graphs.

\section{Evaluated Baselines}
\label{sec:baselines}

To systematically evaluate model performance on the proposed Dual Volatility datasets across both topological evolution (TLP) and semantic drift (TNC), we benchmark 17 representative algorithms spanning multiple modeling paradigms. Based on the type of predictor, we categorize these methods into two groups: \emph{TGNN-Predictors} and \emph{LLM-Predictors}.

\subsection{TGNN-Predictors}

This paradigm relies on Temporal Graph Neural Networks (TGNNs) as the predictive backbone, operating directly on graph structures via message passing or temporal aggregation. Most existing methods specifically designed for temporal graph learning fall into this category, including both purely structural TGNNs and hybrid approaches that incorporate LLMs as auxiliary components.

\textbf{Pure TGNNs.}
As the dominant paradigm in temporal graph learning, pure TGNNs serve as strong structural baselines. We evaluate nine state-of-the-art models covering the major mechanisms for modeling topological evolution. These include memory-based continuous-time models (JODIE~\cite{kumar2019predicting}, DyRep~\cite{trivedi2019dyrep}, TGN~\cite{rossi2020temporal}), attention-based temporal aggregation methods (TGAT~\cite{xu2020inductive}, DyGFormer~\cite{yu2023towards}, FreeDyG~\cite{tian2024freedyg}), and walk-based or hybrid message-fusing approaches (CAWN~\cite{wang2021inductive}, TCL~\cite{wang2021tcl}, GraphMixer~\cite{cong2023we}). These models primarily rely on structural patterns and temporal interaction frequencies to make predictions.

\textbf{LLM-as-Enhancer.}
This hybrid subcategory leverages frozen LLMs to enrich semantic representations while retaining TGNNs as the final predictors. We evaluate two recent state-of-the-art methods: LKD4DyTAG~\cite{roy2025llm} and CROSS~\cite{zhang2025unifying}. LKD4DyTAG prompts an LLM to extract semantic embeddings from nodes’ historical interactions and propagates them along the graph structure, using knowledge distillation from LLMs to guide TGNN optimization. In contrast, CROSS first generates textual summaries of interaction histories via an LLM, and then fuses the resulting embeddings with TGNN representations through multi-layer neural networks. Both approaches aim to enhance semantic awareness without altering the underlying structural modeling mechanism of TGNNs.

\subsection{LLM-Predictors}

An emerging line of work directly employs LLMs as predictors for graph learning, bypassing traditional message-passing frameworks. These methods typically convert graph structures into sequential text or embedding-based prompts, allowing LLMs to act as the primary reasoning engine. Although most existing methods in this paradigm are designed for static graphs, with the exception of the recent TGTalker~\cite{huang2025are}, their flexibility enables adaptation to temporal settings. We therefore extend representative approaches to temporal graphs to evaluate their ability to model both structural and semantic evolution.

\textbf{In-Context Learning (ICL).}
These training-free methods leverage the pretrained knowledge and reasoning capabilities of LLMs, adapting to graph tasks through prompt-based demonstrations. TGTalker belongs to this category, constructing prompts from temporal interaction triplets $(u,v,t)$ to guide the LLM in generating predictions. Building upon this, we evaluate two Disentangled Spatial-Temporal Thought variants from LLM4DyG~\cite{zhang2024llm4dyg}: DST-v1, which encodes structural information before temporal context, and DST-v2, which reverses this order. These methods serialize temporal graphs as linear sequences of triplets, requiring the LLM to infer structural relationships implicitly.

\textbf{Supervised Fine-Tuning (SFT).}
We adapt two recent LLM-as-Predictor methods originally proposed for static graphs, LLaGA~\cite{chen2024llaga} and GraphGPT~\cite{tang2024graphgpt}, to temporal settings. These methods construct prompts that include task instructions and sequences of structure-aware embeddings, and employ a learnable projector to align graph embeddings with the LLM embedding space. For LLaGA, we evaluate both variants: LLaGA-ND, which uses the initial embeddings of the center node and its neighbors, and LLaGA-HO, which uses embeddings obtained after multiple rounds of message passing over historical interaction structure. GraphGPT encodes graph structure through aligned graph and text embeddings, and additionally fine-tunes the LLM embedding layer. Unlike ICL methods, SFT approaches provide explicitly structured and hierarchical representations of graph topology.

Detailed descriptions of all 17 methods, along with prompt templates for LLM-Predictors, are provided in Appendix~\ref{appx:method}.

\section{Experiments and Analysis}
\label{sec:exp}

\paragraph{Implementation Details.}
For TGNN-based methods, we follow the training protocols and pipelines established in DyGLib~\cite{yu2023towards}, first training all models on the TLP task and then using the resulting checkpoints to initialize training for TNC. For all LLM-based methods, we adopt Qwen3-8B~\cite{qwen3technicalreport} as the backbone model. The hyperparameter configurations and training procedures for SFT-based approaches strictly follow their original implementations~\cite{chen2024llaga, tang2024graphgpt}. Sentence-BERT~\cite{reimers-2019-sentence-bert} is used as the default text encoder to extract embeddings for all textual attributes.
For both tasks, each dataset is chronologically split into training, validation, and test sets with a 40\%/10\%/50\% ratio. All experiments are conducted on NVIDIA RTX A6000 GPUs (48GB). Additional implementation details are provided in Appendix~\ref{appx:setup}.

\paragraph{Evaluation Settings and Metrics.}
We follow standard evaluation protocols~\cite{poursafaei2022towards,yu2023towards} by considering both transductive and inductive settings, along with three negative sampling strategies for TLP: random, historical, and inductive. For evaluation metrics, we adopt widely used measures including Average Precision (AP) and Area Under the ROC Curve (AUROC), as well as Mean Reciprocal Rank (MRR), which has gained increasing attention in recent work~\cite{huang2023temporal,yi2025tgbseq}.
Due to space constraints and the consistent trends observed between MRR and the other metrics across different settings, we report AP and AUROC under the transductive setting with random negative sampling in the main text, while deferring MRR results and full experimental details under all settings to Appendix~\ref{appx:more-results}.
For the temporal node classification task, we use Macro-F1 and Balanced Accuracy (bACC) to account for class imbalance, following prior studies~\cite{brodersen2010balanced, grandini2020metrics,he2009learning,hinojosa2024performance,opitz2022bias,tarekegn2021review}.

\subsection{Main Results}
\label{sec:main-res}

\begin{table*}[t]
\caption{Temporal Link Prediction Results. Results are averaged over three independent runs (in \%). The best and second-best results for each dataset are highlighted in \textbf{bold} and \underline{underlined}, respectively.}
\centering
\begin{adjustbox}{width=1.\textwidth}
\begin{tabular}{ccccccccccccc}
\toprule

\multicolumn{1}{c}{\multirow{2}{*}{Methods}} & \multicolumn{2}{c}{FOOD} & \multicolumn{2}{c}{IMDB} & \multicolumn{2}{c}{Librarything} & \multicolumn{2}{c}{Beeradvocate} & \multicolumn{2}{c}{Amazon-Kindle} & \multicolumn{2}{c}{Ratebeer} \\
\multicolumn{1}{c}{} & AP & AUROC & AP &AUROC & AP & AUROC & AP & AUROC & AP & AUROC & AP & AUROC \\
\midrule

EdgeBank & \ms{50.00}{0.00} & \ms{49.70}{0.00} & \ms{50.00}{0.00} & \ms{46.29}{0.00} & \ms{50.00}{0.00} & \ms{49.77}{0.00} & \ms{50.51}{0.00} & \ms{50.57}{0.00} & \ms{50.00}{0.00} & \ms{49.98}{0.00} & \ms{50.07}{0.00} & \ms{49.74}{0.00} \\ 
JODIE & \mstwo{72.96}{1.88} & \mstwo{71.95}{0.69} & \mstwo{60.30}{2.37} & \mstwo{65.74}{3.20} & \mstwo{75.76}{0.63} & \mstwo{73.62}{0.75} & \ms{79.75}{11.63} & \ms{80.93}{9.53} & \mstwo{82.81}{2.00} & \ms{81.86}{1.69} & \msone{92.62}{0.18} & \msone{91.56}{0.36} \\ 
DyRep & \ms{72.83}{0.88} & \ms{69.01}{1.53} & \msone{64.79}{1.30} & \msone{69.79}{2.03} & \ms{73.81}{1.55} & \ms{72.34}{1.13} & \msone{89.61}{0.20} & \msone{87.01}{0.40} & \ms{78.35}{2.63} & \ms{76.63}{2.69} & \ms{85.87}{0.33} & \ms{83.84}{1.18} \\ 
TGAT & \ms{54.79}{1.25} & \ms{54.38}{0.67} & \ms{54.55}{0.01} & \ms{57.85}{0.05} & \ms{64.20}{0.89} & \ms{65.68}{0.96} & \ms{59.37}{3.59} & \ms{66.58}{2.13} & \ms{79.00}{1.97} & \ms{81.12}{1.32} & \ms{58.93}{4.20} & \ms{64.38}{2.29} \\ 
TGN & \msone{77.74}{0.05} & \msone{75.74}{0.28} & \ms{45.40}{0.89} & \ms{48.79}{1.66} & \msone{82.60}{0.22} & \msone{82.02}{0.09} & \ms{83.46}{1.68} & \ms{83.25}{1.06} & \ms{71.77}{2.90} & \ms{79.92}{1.49} & \mstwo{91.55}{0.24} & \mstwo{91.41}{0.24} \\ 
CAWN & \ms{51.71}{0.38} & \ms{53.96}{0.23} & \ms{54.00}{1.17} & \ms{55.60}{2.75} & \ms{62.01}{0.21} & \ms{63.21}{0.07} & \ms{66.15}{0.85} & \ms{71.29}{0.05} & \ms{79.34}{1.50} & \ms{80.79}{1.22} & \ms{59.94}{2.40} & \ms{65.78}{0.67} \\ 
TCL & \ms{49.12}{0.79} & \ms{50.98}{1.23} & \ms{55.78}{0.49} & \ms{58.83}{0.44} & \ms{54.51}{4.02} & \ms{56.63}{2.91} & \ms{55.40}{0.18} & \ms{64.62}{0.29} & \ms{79.68}{1.51} & \ms{80.96}{0.95} & \ms{57.61}{3.77} & \ms{62.63}{2.39} \\ 
GraphMixer & \ms{66.98}{1.71} & \ms{65.78}{0.93} & \ms{57.01}{0.03} & \ms{61.60}{0.09} & \ms{62.13}{1.45} & \ms{63.16}{2.44} & \mstwo{86.02}{0.24} & \mstwo{85.57}{0.29} & \ms{81.53}{0.28} & \mstwo{81.87}{0.09} & \ms{78.53}{4.22} & \ms{78.05}{4.06} \\ 
DyGFormer & \ms{59.48}{0.24} & \ms{57.26}{0.52} & \ms{55.32}{0.11} & \ms{58.42}{0.10} & \ms{54.53}{2.16} & \ms{54.35}{3.14} & \ms{50.53}{0.08} & \ms{56.80}{0.05} & \ms{72.78}{0.32} & \ms{74.71}{0.09} & \ms{53.28}{0.91} & \ms{57.65}{0.73} \\ 
FreeDyG & \ms{72.40}{0.34} & \ms{69.60}{0.37} & \ms{57.53}{0.89} & \ms{62.16}{0.93} & \ms{66.08}{0.26} & \ms{65.13}{1.70} & \ms{82.93}{6.75} & \ms{85.18}{4.07} & \ms{67.26}{3.83} & \ms{71.25}{1.99} & \ms{69.49}{3.30} & \ms{74.16}{1.86} \\ 
LKD4DyTAG & \ms{66.63}{0.95} & \ms{63.62}{0.95} & \ms{56.04}{0.92} & \ms{59.13}{0.53} & \ms{59.94}{0.49} & \ms{62.17}{0.34} & \ms{79.42}{1.15} & \ms{79.52}{0.91} & \ms{80.64}{0.46} & \ms{81.64}{0.42} & \ms{69.51}{0.03} & \ms{70.85}{0.05} \\ 
CROSS & \ms{58.15}{0.13} & \ms{56.48}{0.24} & \ms{55.98}{0.05} & \ms{59.14}{0.12} & \ms{59.04}{0.83} & \ms{60.25}{0.94} & \ms{51.36}{1.03} & \ms{57.52}{1.01} & \msone{87.25}{0.13} & \msone{88.31}{0.12} & \ms{52.57}{0.09} & \ms{57.19}{0.21} \\ 
\cmidrule{1-13}
TGTalker & \ms{51.04}{0.02} & \ms{52.02}{0.03} & \ms{50.22}{0.04} & \ms{50.43}{0.07} & \ms{50.63}{ 0.01} & \ms{51.24}{0.04} & \ms{51.18}{0.03} & \ms{51.36}{0.02} & \ms{52.54}{0.09} & \ms{54.79}{0.11} & \ms{51.11}{0.07} & \ms{52.15}{0.06} \\ 
DST-v1 & \ms{51.31}{0.01} & \ms{52.53}{0.03} & \ms{50.26}{0.10} & \ms{50.52}{0.20} & \ms{50.71}{0.02} & \ms{51.39}{0.06} & \ms{51.34}{0.04} & \ms{51.47}{0.05} & \ms{52.66}{0.10} & \ms{54.85}{0.09} & \ms{51.23}{0.04} & \ms{51.26}{0.03} \\ 
DST-v2 & \ms{51.33}{0.00} & \ms{52.57}{0.01} & \ms{50.35}{0.10} & \ms{50.70}{0.19} & \ms{50.61}{0.01} & \ms{51.19}{0.02} & \ms{51.21}{0.01} & \ms{51.53}{0.01} & \ms{52.62}{0.08} & \ms{54.81}{0.08} & \ms{51.05}{0.01} & \ms{51.07}{0.02} \\ 
Llaga-ND & \ms{53.10}{0.36} & \ms{55.52}{0.57} & \ms{52.01}{0.75} & \ms{53.55}{1.19} & \ms{53.98}{0.07} & \ms{53.95}{0.14} & \ms{54.97}{0.03} & \ms{54.94}{0.06} & \ms{68.62}{3.37} & \ms{72.85}{3.58} & \ms{54.00}{0.02} & \ms{54.99}{0.05} \\ 
Llaga-HO & \ms{61.95}{1.32} & \ms{65.67}{1.47} & \ms{54.37}{0.82} & \ms{57.34}{1.06} & \ms{56.87}{0.02} & \ms{56.88}{0.03} & \ms{58.00}{0.05} & \ms{58.00}{0.10} & \ms{72.64}{1.11} & \ms{76.63}{1.54} & \ms{56.09}{0.07} & \ms{56.15}{0.11} \\ 
GraphGPT & \ms{55.09}{0.43} & \ms{57.95}{1.02} & \ms{51.80}{0.18} & \ms{53.19}{0.29} & \ms{54.96}{0.04} & \ms{54.89}{0.11} & \ms{56.80}{6.80} & \ms{58.35}{8.35} & \ms{70.72}{0.72} & \ms{71.17}{1.13} & \ms{56.03}{6.03} & \ms{59.29}{9.29} \\   	 	

\bottomrule
\end{tabular}
\end{adjustbox}
\label{tab:main-lp}
\end{table*}

From the results in \cref{tab:main-lp,tab:main-nc}, we derive several key observations.

\textbf{TGNNs and LLM-Predictors exhibit a clear capability divide.}
TGNN-based predictors achieve strong performance on TLP but degrade substantially on TNC, whereas LLM-based predictors excel on TNC while struggling on TLP. Since TLP primarily requires modeling temporal structural evolution and TNC emphasizes semantic drift, this divergence suggests that each paradigm captures only one facet of temporal information. These results reveal a fundamental limitation of existing approaches in achieving balanced structural and semantic learning.

\textbf{Naive memorization fails under high novelty.}
The pure memorization baseline, EdgeBank, performs close to random (AUROC $\approx 50$) across all datasets, confirming its ineffectiveness in high-novelty scenarios. This underscores the necessity of models that can learn evolving temporal patterns rather than relying on historical repetition.

\textbf{Continuous state tracking is critical for TLP under structural volatility.}
Memory-based TGNNs, including TGN, JODIE, and DyRep, consistently rank among the strongest methods on TLP. By maintaining evolving node states, these models capture cumulative temporal effects, which appears to be a key factor underlying their superior performance when interaction repetition is limited.

\textbf{Explicit structural inputs help LLM-Predictors on TLP, but the learning paradigm remains the bottleneck.}
\Cref{tab:main-lp} reveals three key findings. (i) Linearized temporal-structural sequences in ICL inputs, regardless of ordering, perform close to random, indicating that prompting off-the-shelf LLMs is insufficient for extracting implicit structural signals from simple $(u,v,t)$ serializations. (ii) SFT consistently outperforms ICL, suggesting that incorporating explicit structural encoding in the input benefits TLP performance. (iii) Within SFT-based methods, LLaGA-HO and GraphGPT outperform LLaGA-ND, demonstrating that structure-aware aggregation of node embeddings partially compensates for LLMs' limited structural understanding. Nevertheless, even the best-performing LLM-Predictors on TLP remain significantly behind TGNNs, indicating that input-level structural enhancements alone are insufficient and that the limitation lies in the underlying learning paradigm.

\textbf{LLM-as-Enhancer struggles to address TGNN semantic limitations.}
LLM-enhanced TGNN methods such as CROSS and LKD4DyTAG achieve competitive performance on TLP but inherit the weaknesses of TGNNs on TNC. This suggests that simply injecting LLM-derived semantic information into an aggregation framework is insufficient to overcome the structural bias of TGNNs, limiting their ability to model rapidly evolving semantics.

\begin{table*}[t]
\caption{Temporal Node Classification Results. Results are averaged over three independent runs (in \%). The best and second-best results for each dataset are highlighted in \textbf{bold} and \underline{underlined}, respectively.}
\centering
\begin{adjustbox}{width=0.98\textwidth}
\begin{tabular}{ccccccccccc}
\toprule

\multicolumn{1}{c}{\multirow{2}{*}{Methods}} & \multicolumn{2}{c}{FOOD} & \multicolumn{2}{c}{IMDB} & \multicolumn{2}{c}{Beeradvocate} & \multicolumn{2}{c}{Amazon-Kindle} & \multicolumn{2}{c}{Ratebeer} \\

\multicolumn{1}{c}{} & Macro-F1 & mACC & Macro-F1 & mACC & Macro-F1 & mACC & Macro-F1 & mACC & Macro-F1 & mACC \\
\midrule

JODIE & \ms{25.95}{4.66} & \ms{33.48}{10.52} & \ms{12.65}{0.90} & \ms{15.85}{0.57} & \ms{7.00}{0.06} & \ms{14.52}{0.15} & \ms{17.60}{0.01} & \ms{13.64}{0.04} & \ms{7.86}{0.30} & \ms{13.88}{0.32} \\ 
DyRep & \ms{21.22}{0.10} & \ms{22.93}{0.10} & \ms{13.74}{0.82} & \ms{15.97}{0.68} & \ms{6.50}{0.34} & \ms{14.28}{0.01} & \ms{13.36}{1.43} & \ms{13.51}{0.89} & \ms{7.38}{1.33} & \ms{13.58}{0.98} \\ 
TGAT & \ms{22.73}{2.73} & \ms{27.78}{5.56} & \ms{8.53}{0.02} & \ms{12.48}{0.00} & \ms{6.18}{0.04} & \ms{14.30}{0.01} & \ms{14.91}{0.37} & \ms{14.38}{0.25} & \ms{8.67}{0.05} & \ms{14.53}{0.06} \\ 
TGN & \ms{21.98}{1.20} & \ms{23.61}{0.96} & \ms{8.51}{0.00} & \ms{12.50}{0.00} & \ms{8.35}{0.01} & \ms{15.31}{0.02} & \ms{13.22}{0.25} & \ms{13.28}{0.19} & \ms{7.34}{1.15} & \ms{13.54}{0.78} \\ 
CAWN & \ms{20.28}{0.29} & \ms{22.38}{0.15} & \ms{11.48}{2.98} & \ms{18.75}{6.25} & \ms{8.77}{0.27} & \ms{15.42}{0.02} & \ms{14.96}{0.33} & \ms{14.43}{0.25} & \ms{8.79}{0.04} & \ms{14.53}{0.08} \\ 
TCL & \ms{19.99}{0.02} & \ms{22.22}{0.01} & \ms{14.56}{1.19} & \ms{15.46}{0.04} & \ms{9.26}{0.28} & \ms{16.03}{0.33} & \ms{24.25}{0.68} & \ms{21.76}{0.75} & \ms{14.34}{0.67} & \ms{19.35}{0.96} \\ 
GraphMixer & \ms{16.94}{0.01} & \ms{20.77}{0.02} & \ms{9.20}{0.69} & \ms{12.91}{0.41} & \ms{6.14}{0.01} & \ms{14.29}{0.01} & \ms{13.35}{0.40} & \ms{13.38}{0.22} & \ms{7.13}{1.78} & \ms{13.59}{1.09} \\ 
DyGFormer & \ms{36.58}{3.02} & \ms{36.39}{3.92} & \ms{20.05}{11.54} & \ms{20.01}{7.51} & \ms{12.91}{1.59} & \ms{18.93}{0.48} & \ms{20.19}{0.60} & \ms{18.14}{0.43} & \ms{15.52}{1.37} & \ms{18.42}{1.12} \\ 
FreeDyG & \ms{20.46}{0.04} & \ms{22.47}{0.02} & \ms{10.20}{1.55} & \ms{13.46}{0.89} & \ms{7.75}{1.61} & \ms{14.91}{0.62} & \ms{12.80}{0.86} & \ms{13.10}{0.47} & \ms{8.54}{0.46} & \ms{14.23}{0.35} \\ 
LKD4DyTAG & \ms{23.37}{3.38} & \ms{27.78}{5.56} & \ms{8.51}{0.01} & \ms{12.50}{0.02} & \ms{6.14}{0.01} & \ms{14.29}{0.02} & \ms{10.39}{ 0.01} & \ms{11.11}{0.04} & \ms{5.80}{0.07} & \ms{12.57}{0.01} \\ 
CROSS & \ms{19.99}{0.02} & \ms{22.22}{0.01} & \ms{8.62}{0.02} & \ms{12.41}{0.01} & \ms{6.17}{0.01} & \ms{14.30}{0.01} & \ms{9.59}{0.07} & \ms{10.97}{0.03} & \ms{5.35}{0.00} & \ms{12.50}{0.00} \\ 
\cmidrule{1-11}
TGTalker & \ms{31.84}{0.05} & \ms{30.13}{0.52} & \ms{28.96}{0.59} & \ms{30.63}{0.46} & \msone{16.52}{0.31} & \ms{18.00}{0.30} & \mstwo{25.60}{0.24} & \ms{24.02}{0.26} & \ms{13.42}{0.47} & \ms{15.29}{0.78} \\ 
DST-v1 & \ms{32.87}{0.22} & \ms{31.62}{0.26} & \msone{29.37}{0.33} & \mstwo{30.94}{0.69} & \ms{16.29}{0.42} & \ms{17.47}{0.38} & \msone{26.00}{0.33} & \mstwo{24.42}{0.14} & \ms{12.83}{0.92} & \ms{14.89}{0.65} \\ 
DST-v2 & \msone{37.59}{0.25} & \msone{39.91}{0.22} & \mstwo{29.14}{0.24} & \ms{30.72}{0.36} & \ms{16.12}{0.25} & \ms{17.17}{0.34} & \ms{24.64}{0.37} & \ms{23.94}{0.25} & \ms{13.34}{0.94} & \ms{14.69}{0.50} \\ 
Llaga-ND & \ms{28.48}{0.30} & \ms{32.04}{0.66} & \ms{19.13}{1.13} & \ms{25.95}{1.16} & \ms{13.90}{1.88} & \ms{18.30}{1.60} & \ms{18.57}{2.53} & \ms{17.25}{1.13} & \ms{14.12}{0.08} & \ms{17.51}{1.19} \\ 
Llaga-HO & \ms{30.37}{0.41} & \ms{33.26}{1.75} & \ms{20.66}{1.63} & \ms{26.45}{1.67} & \ms{15.09}{1.70} & \mstwo{19.24}{1.35} & \ms{21.01}{2.68} & \ms{18.66}{2.55} & \mstwo{16.94}{0.17} & \msone{20.63}{1.42} \\ 
GraphGPT & \mstwo{36.74}{2.83} & \mstwo{38.40}{1.47} & \ms{25.79}{5.59} & \msone{41.01}{16.08} & \mstwo{16.31}{0.28} & \msone{20.22}{0.37} & \ms{24.28}{4.63} & \msone{24.74}{0.16} & \msone{18.93}{0.16} & \mstwo{20.21}{0.20} \\ 
 
\bottomrule
\end{tabular}
\end{adjustbox}
\label{tab:main-nc}
\end{table*}

\subsection{Why Do TGNNs Fail at Semantic Tracking?}
\label{subsec:analysis}

\begin{wrapfigure}{r}{0.5\textwidth}
    \centering
    \includegraphics[width=\linewidth]{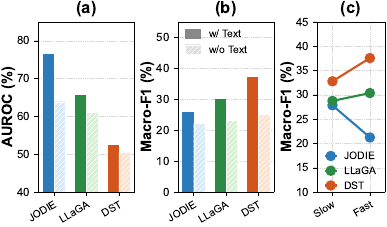}
    \caption{Comparative analysis of representative TGNN and LLM-Predictor methods.
(a) and (b) show the performance of each method on TLP and TNC, respectively, with and without textual inputs.
(c) illustrates how model performance varies as the frequency of semantic changes increases.}
    \label{fig:analysis}
\end{wrapfigure}

To systematically analyze why TGNNs fail catastrophically on semantic tracking (TNC), we conduct in-depth probing studies to isolate their underlying bottlenecks. We select JODIE as a representative TGNN due to its consistently strong performance, and choose DST-v2 and LLaGA-HO—two top-performing LLM-Predictors on TNC—as representatives of ICL- and SFT-based methods, respectively. The comparative analysis (Figure~\ref{fig:analysis}) reveals several fundamental limitations of TGNNs.

\textbf{Architectural semantic blindness.}
We first examine whether models effectively utilize textual attributes for semantic tracking. As shown in Figure~\ref{fig:analysis}(a) and (b), removing textual attributes causes a sharp drop in TLP performance for TGNNs but a much smaller impact on TNC, whereas LLM-Predictors suffer substantial degradation on TNC with minor impact on TLP. This consistent trend holds across all methods in both paradigms (see \cref{fig:flp-text} for full results). The observed asymmetry suggests that, although TGNNs can exploit text as auxiliary signals for structural modeling, their message-aggregation architectures are inherently limited in leveraging textual information for semantic tracking. 

\textbf{Representation lock-in.}
We next investigate whether this limitation can be mitigated through downstream supervision. Specifically, we compare two training strategies for TGNNs on TNC: linear probing (training only the classifier head on a TLP-pretrained backbone) and end-to-end fine-tuning (jointly optimizing the backbone and classifier). Surprisingly, end-to-end fine-tuning yields no meaningful improvement over linear probing and can even degrade performance (\cref{tab:lp-e2e}). This suggests a strong \emph{representation lock-in}, where gradients from semantic supervision are insufficient to reshape representations that are already heavily biased toward structural patterns. As a result, the learned embeddings remain anchored in topological dynamics and are not adaptable to semantic drift.

\textbf{Structural inertia under semantic volatility.}
Finally, we analyze model performance under varying degrees of semantic change based on the frequency of label drifts. As shown in Figure~\ref{fig:analysis}(c), TGNN performance deteriorates sharply as the environment transitions from slow (1–2 changes) to fast (>2 changes) semantic drift. In contrast, LLM-Predictors maintain stable or even improved performance. This highlights an inherent \emph{structural inertia} in TGNNs: their reliance on aggregating historical interactions becomes a liability when semantic states change rapidly. Conversely, LLM-Predictors exhibit strong semantic adaptability, dynamically leveraging immediate textual context to track evolving semantics without being constrained by historical structural dependencies.


\subsection{Efficiency Analysis}
\label{subsec:efficiency}

\begin{figure}[t]
  \centering
  \includegraphics[width=0.98\linewidth]{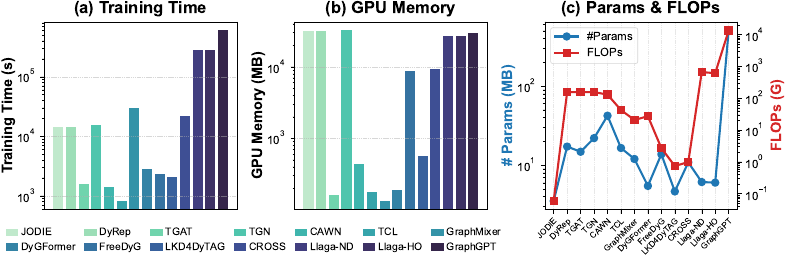}
  \caption{Efficiency Results. Scalability results of all methods are shown in~\cref{fig:scalability}.}
  \label{fig:effiency}
\end{figure}

We evaluate the practical applicability of all benchmarked methods by analyzing their computational efficiency. \cref{fig:effiency} reports training time, GPU memory consumption, parameter counts, and FLOPs for all methods on the largest-scale Amazon-Kindle dataset.

\textbf{Staggering overhead of LLM-Predictor paradigms.}
The most striking observation is the substantial computational cost incurred by LLM-Predictor paradigms due to the use of generative LLMs. In particular, GraphGPT—whose training involves fine-tuning the embedding layer of an LLM—requires nearly 600{,}000 seconds of training time and 13.5 teraFLOPs, exceeding even the most resource-intensive TGNNs by several orders of magnitude. Although LLM-Predictors achieve strong performance on semantic tracking (TNC), their prohibitive computational overhead limits their practicality in real-world applications, highlighting the need for fundamental improvements to make this paradigm deployable in realistic settings.

\textbf{Performance–efficiency trade-off within TGNNs.}
Clear trade-offs are observed among TGNNs. Memory-based architectures (e.g., JODIE, DyRep, and TGN) achieve dominant performance on TLP, but incur substantially higher computational and memory overhead due to the need to maintain and continuously update recurrent node states. This reflects the inherent cost of continuous-time structural tracking required to handle structural volatility effectively.

\textbf{Scalability robustness.}
Beyond static efficiency, we further evaluate scalability by measuring runtime on progressively larger subsets of the Amazon-Kindle dataset, ranging from 1M to 5M interactions. Results show that all methods exhibit approximately linear scaling with respect to dataset size, indicating predictable and robust scalability. Detailed results are provided in \cref{fig:scalability}.

\section{Conclusions}

In this work, we introduce TTGBench, a benchmark for jointly evaluating structural and semantic evolution in temporal graphs. By constructing six text-rich datasets with \emph{dual volatility}, TTGBench provides a challenging and realistic setting that requires modeling both structural novelty and semantic dynamics. Extensive experiments on 17 state-of-the-art methods reveal a clear \emph{capability divide}: TGNN-based methods excel at structural modeling but struggle with semantic tracking, whereas LLM-based predictors exhibit the opposite trend. Through in-depth analysis, we attribute this phenomenon to intrinsic architectural biases in the two paradigms and further identify clear trade-offs between performance and efficiency in existing approaches.

Our findings suggest several promising directions for future research. A key challenge is to develop unified models that can simultaneously capture structural and semantic dynamics. In addition, improving structural reasoning in LLMs and enhancing the semantic adaptability of TGNNs are both critical for bridging the current gap. Finally, designing efficient and scalable architectures remains essential for practical deployment.


\bibliographystyle{plain}
\bibliography{paper}


\appendix
\clearpage
\appendix

\section*{Appendix}

\section{Datasets Details}
\label{appx:dataset}

\subsection{Dataset Description}

\begin{table*}[!h]
    \centering
    \caption{Comparison between our datasets and existing temporal graph datasets.}
    \label{tab:cmp-datasets}
    \begin{adjustbox}{width=1.\textwidth}
    \begin{tabular}{c|crrrrcccc}
        \toprule
        & Dataset & Nodes & Total Edges & Unique Edges & Unique Steps & Text Attr & Novelty & Surprise & Node Labels \\
        \midrule
        \multirow{15}{*}{Previous}
            & mooc & 7,144 & 411,749 & 178,443 & 345,600 & \xmark & 0.433 & 0.785     & N.A.  \\ 
            & lastfm & 1,980 & 1,293,103 & 154,993 & 1,283,614 & \xmark & 0.12 & 0.369  & N.A. \\ 
            & enron & 184 & 125,235 & 3,125 & 22,632 & \xmark & 0.076 & 0.402        & N.A. \\ 
            & SocialEvo & 74 & 2,099,519 & 4,486 & 565,932 & \xmark & 0.002 & 0.027   & N.A. \\ 
            & uci & 1,899 & 59,835 & 20,296 & 58,911 & \xmark & 0.339 & 0.796         & N.A. \\ 
            & Flights & 13,169 & 1,927,145 & 395,072 & 122 & \xmark & 0.194 & 0.362   & N.A. \\ 
            & CanParl & 734 & 74,478 & 51,331 & 14 & \xmark & 0.673 & 0.654         & N.A. \\ 
            & USLegis & 225 & 60,396 & 26,423 & 12 & \xmark & 0.437 & 0.340          & N.A. \\ 
            & UNtrade & 255 & 507,497 & 36,182 & 32 & \xmark & 0.09 & 0.051         & N.A. \\ 
            & UNvote & 201 & 1,035,742 & 31,516 & 72 & \xmark & 0.056 & 0.017        & N.A. \\ 
            & Contacts & 692 & 2,426,279 & 79,530 & 8,064 & \xmark & 0.023 & 0.291    & N.A. \\ 
            & tgbl-wiki & 9,227 & 157,474 & 18,257 & 152,757 & \xmark & 0.116 & 0.108 & N.A. \\ 
            & tgbn-reddit & 11,766 & 27,174,118 & 516,669 & 21,889,537 & \xmark & 0.02 & 0.013 & N.A.\\  
            & GDELT & 6,786 & 1,339,245 & 249,241 & 2,403 & \cmark & 0.22 & 0.562 & N.A. \\ 
            & ICEWS1819 & 31,796 & 1,100,071 & 314,011 & 730 & \cmark & 0.292 & 0.695 & N.A. \\ 
        \midrule
        \multirow{6}{*}{\textbf{Ours}} 
            & FOOD & 33,773 & 401,057 & 401,057 & 6,252 & \cmark & 1.0 & 1.0  & 9  \\ 
            & IMDB & 32,371 & 310,891 & 310,891 & 7,099 & \cmark & 1.0 & 1.0  & 8 \\ 
            & Librarything & 51,201 & 785,690 & 785,690 & 2,914 & \cmark & 1.0 & 1.0  & N.A. \\ 
            & Beeradvocate & 99,361 & 1,586,573 & 1,571,767 & 1,577,920 & \cmark & 0.991 & 0.997 & 7 \\ 
            & Ratebeer & 139,538 & 2,924,105 & 2,855,175 & 4,254 & \cmark & 1.0 & 1.0  & 8 \\ 
            & Amazon-Kindle & 215,504 & 5,621,343 & 5,561,738 & 5,484,604 & \cmark & 1.0 & 1.0  & 8 \\ 
        \bottomrule
    \end{tabular}
    \end{adjustbox}
\end{table*}

The detailed descriptions of the six proposed datasets are provided below, with comparisons to existing datasets summarized in~\cref{tab:cmp-datasets}.

\begin{itemize}[leftmargin=*]
\item \textbf{FOOD}~\cite{majumder2019generating}
This dataset contains recipes and user reviews collected from Food.com (formerly GeniusKitchen), covering 18 years of temporal interactions. It provides rich signals for studying culinary trends, evolving user preferences, and long-term temporal dynamics. In the constructed graph, users and recipes are represented as nodes, and a timestamped edge is created when a user reviews a recipe at time $t$. Each recipe contains detailed cooking instructions, ingredient information, and category labels. We assign recipe categories as users’ temporal interest labels, representing users’ evolving interests at each interaction. Since each recipe may have multiple labels, FOOD supports \textit{multi-label} temporal node classification.
\item \textbf{IMDB}~\cite{misra2019imdb}
This dataset is constructed from IMDB and consists of users, movies, and timestamped review interactions. Each movie includes plot summaries and genre tags, while each review contains textual content and temporal information. We use movie genre tags as users’ temporal interest labels at the time of interaction, making IMDB another \textit{multi-label} classification benchmark. The rich textual information in both movies and reviews provides abundant signals for modeling both structural evolution and semantic drift in temporal graphs.
\item \textbf{LibraryThing}~\cite{cai2017spmc,zhao2015improving}
Collected from the book review platform LibraryThing (\url{https://www.librarything.com/}
), this dataset contains users, books, and timestamped textual review interactions. It captures diverse literary preferences and their evolution over time. In our temporal graph, users and books are nodes connected through review-based temporal edges. Since this dataset does not provide item category labels, it is used only for temporal link prediction.
\item \textbf{BeerAdvocate}~\cite{mcauley2012learning}
This dataset consists of beer reviews involving users and beers as graph nodes. Each beer is associated with a unique style label, which we use as the user’s temporal interest label during interaction. Because each beer has a single category label, BeerAdvocate supports \textit{multi-class} temporal node classification.
\item \textbf{Amazon-Kindle}~\cite{hou2024bridging}
This dataset is drawn from the Amazon Reviews 2023 collection (\url{https://amazon-reviews-2023.github.io/}
) and focuses on user reviews of Kindle products. Users and products form the graph nodes, while products and reviews both provide rich textual information. The resulting product-review network spans from 1996 to September 2023. Each product has a specific category label, making this dataset suitable for \textit{multi-class} temporal node classification.
\item \textbf{Ratebeer}~\cite{mcauley2013amateurs}
Ratebeer is another large-scale beer review dataset similar to BeerAdvocate, but with much denser temporal interactions, providing a complementary dynamic scenario for temporal graph modeling. Each beer also has a single category label, enabling \textit{multi-class} temporal node classification.
\end{itemize}

\subsection{Dataset Analysis}

\begin{figure*}[!t]
  \centering
  \includegraphics[width=0.92\linewidth]{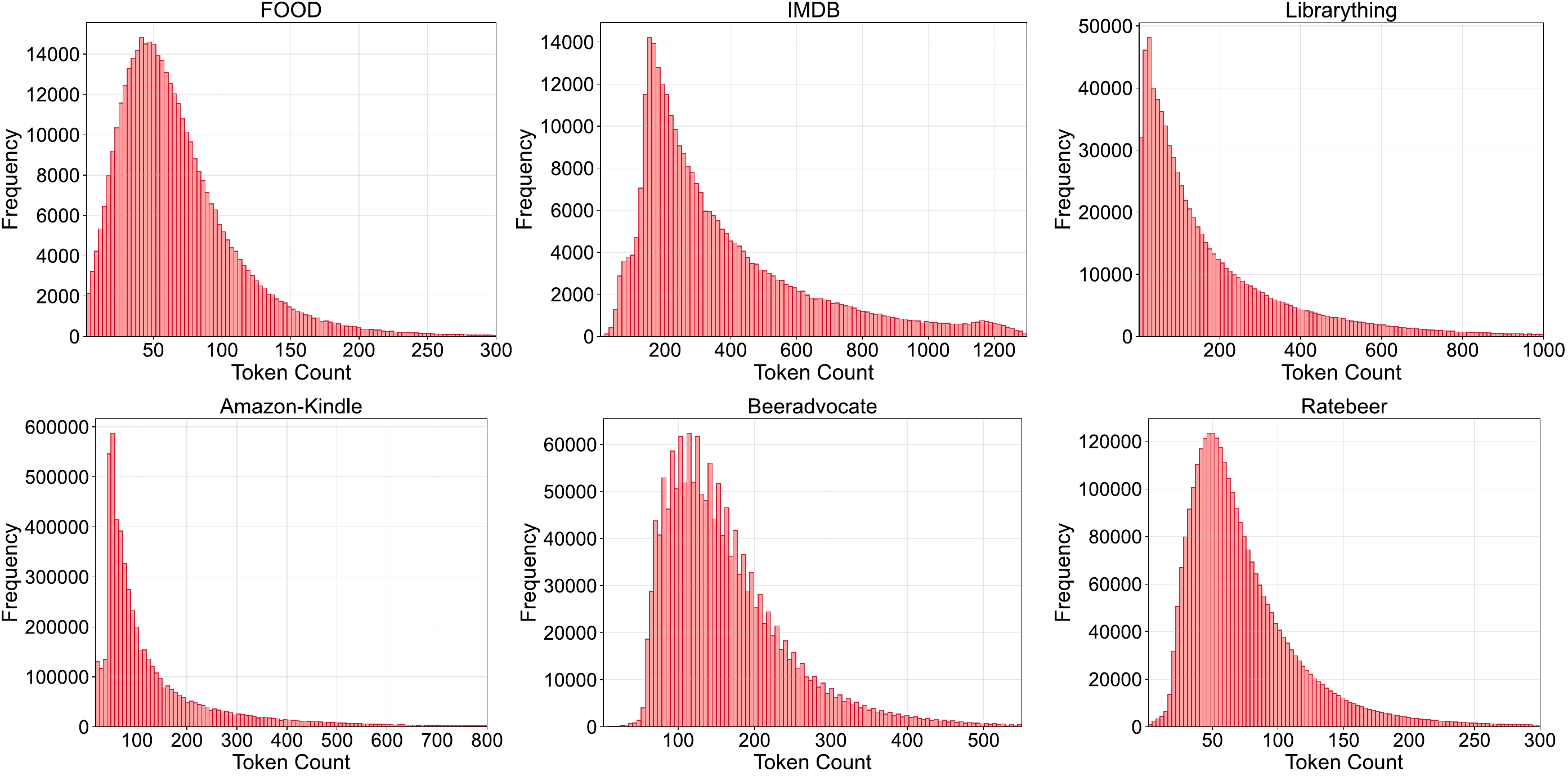}
  \caption{Distribution of edge text length (\#tokens) on TTGBench datasets.}
  \label{fig:token-len}
\end{figure*}

\textbf{Distribution of Edge Text Lengths.} Given that the temporal structure formed by sequential edges is crucial for learning in temporal graph models, we present the distribution of text lengths on edges (measured in tokens) across all datasets in Figure~\ref{fig:token-len}. As shown in the figure, our datasets exhibit a right-skewed distribution, which is a typical characteristic of user-generated content (UGC) in real-world scenarios. Most user interactions consist of brief feedback (e.g., a ``5-star'' rating or a short comment like ``Good''), while only a small fraction of highly engaged users contribute long-tail texts. Additionally, the text length distributions vary across different datasets. These characteristics pose several challenges for temporal graph learning:
\begin{itemize}[leftmargin=*]
    \item The sparsity of information in short texts requires models to capture information over extended temporal windows to effectively associate with structural evolution.
    \item Long texts often contain multiple semantic cues (e.g., multi-faceted product descriptions), demanding that models identify the structure-relevant parts. Moreover, time-sensitive information embedded in long texts adds complexity for precise temporal alignment.
    \item The differing centers of text length distributions across datasets reflect domain-specific language habits, requiring models to adapt to domain variations. Notably, the existence of long-tail distributions also demands robustness from the models.
\end{itemize}


\textbf{Node Label Distribution.}
As shown in Table~\ref{tab:text-labels}, label distributions vary across datasets, with several exhibiting substantial class imbalance. As discussed in~\cref{sec:exp}, we therefore adopt multiple imbalance-aware evaluation metrics for temporal node classification.

\begin{table*}[!t]
    \footnotesize
    \caption{\small Text labels of each dataset.}
    \label{tab:text-labels}
    \centering
    \begin{adjustbox}{width=0.92\textwidth}
    \begin{tabular}{lrl}
        \toprule
        {\cellcolor{white}} Datasets & Text Labels & Percentages  \\
        \midrule
        \multirow{9}{*}{FOOD} & {\cellcolor{white}} Side Dishes \& Vegetables & {\cellcolor{white}} \pcb{10.3}\% \\ 
        ~ & {\cellcolor[HTML]{F2F2F2}} Dietary \& Specialty & {\cellcolor[HTML]{F2F2F2}} \pcb{32.4}\% \\ 
        ~ & {\cellcolor{white}} Soups \& Stews & {\cellcolor{white}} \pcb{1.8}\% \\ 
        ~ & {\cellcolor[HTML]{F2F2F2}} Cuisine \& Regional & {\cellcolor[HTML]{F2F2F2}} \pcb{14.4}\% \\ 
        ~ & {\cellcolor{white}} Desserts \& Sweets & {\cellcolor{white}} \pcb{8.0}\% \\ 
        ~ & {\cellcolor[HTML]{F2F2F2}} Baking \& Breads  & {\cellcolor[HTML]{F2F2F2}} \pcb{3.6}\% \\ 
        ~ & {\cellcolor{white}} Beverages \& Snacks & {\cellcolor{white}} \pcb{5.3}\% \\ 
        ~ & {\cellcolor[HTML]{F2F2F2}} Sauces \& Condiments & {\cellcolor[HTML]{F2F2F2}} \pcb{2.4}\% \\ 
        ~ & {\cellcolor{white}} Main Dishes & {\cellcolor{white}} \pcb{21.8}\% \\ 
        \midrule
        \multirow{8}{*}{IMDB} & {\cellcolor{white}} Horror & {\cellcolor{white}} \pcb{3.7}\% \\ 
        ~ & {\cellcolor[HTML]{F2F2F2}} Family \& Animation & {\cellcolor[HTML]{F2F2F2}} \pcb{4.9}\% \\ 
        ~ & {\cellcolor{white}} Sci-Fi \& Fantasy & {\cellcolor{white}} \pcb{13.0}\% \\ 
        ~ & {\cellcolor[HTML]{F2F2F2}} Drama \& Realism & {\cellcolor[HTML]{F2F2F2}} \pcb{25.0}\% \\ 
        ~ & {\cellcolor{white}} Action \& Adventure & {\cellcolor{white}} \pcb{21.5}\% \\ 
        ~ & {\cellcolor[HTML]{F2F2F2}} Thriller \& Suspense  & {\cellcolor[HTML]{F2F2F2}} \pcb{16.9}\% \\ 
        ~ & {\cellcolor{white}} Historical \& War & {\cellcolor{white}} \pcb{2.7}\% \\ 
        ~ & {\cellcolor[HTML]{F2F2F2}} Comedy \& Lighthearted  & {\cellcolor[HTML]{F2F2F2}} \pcb{12.4}\% \\ 
        \midrule
        \multirow{7}{*}{Beeradvocate} & {\cellcolor{white}} Dark Beers  & {\cellcolor{white}} \pcb{21.9}\% \\ 
        ~ & {\cellcolor[HTML]{F2F2F2}} Strong Ales \& Barley Wines & {\cellcolor[HTML]{F2F2F2}} \pcb{9.2}\% \\ 
        ~ & {\cellcolor{white}} Belgian \& Specialty Ales & {\cellcolor{white}} \pcb{16.5}\% \\ 
        ~ & {\cellcolor[HTML]{F2F2F2}} Sour \& Fruit Beers  & {\cellcolor[HTML]{F2F2F2}} \pcb{5.4}\% \\ 
        ~ & {\cellcolor{white}} Wheat Beers & {\cellcolor{white}} \pcb{6.5}\% \\ 
        ~ & {\cellcolor[HTML]{F2F2F2}} Pale Ales \& IPAs & {\cellcolor[HTML]{F2F2F2}} \pcb{26.3}\% \\ 
        ~ & {\cellcolor{white}} Lagers & {\cellcolor{white}} \pcb{14.3}\% \\ 
        \midrule
        \multirow{8}{*}{Amazon-Kindle} & {\cellcolor{white}} Science, Technology \& Professional & {\cellcolor{white}} \pcb{0.2}\% \\ 
        ~ & {\cellcolor[HTML]{F2F2F2}} Children \& Young Adult & {\cellcolor[HTML]{F2F2F2}} \pcb{6.4}\% \\ 
        ~ & {\cellcolor{white}} Social Sciences \& Humanities & {\cellcolor{white}} \pcb{4.7}\% \\ 
        ~ & {\cellcolor[HTML]{F2F2F2}} Lifestyle \& Hobbies & {\cellcolor[HTML]{F2F2F2}} \pcb{0.5}\% \\ 
        ~ & {\cellcolor{white}} Entertainment \& Arts & {\cellcolor{white}} \pcb{0.8}\% \\ 
        ~ & {\cellcolor[HTML]{F2F2F2}} Nonfiction \& Self-Improvement & {\cellcolor[HTML]{F2F2F2}} \pcb{1.3}\% \\ 
        ~ & {\cellcolor{white}} Short Reads  & {\cellcolor{white}} \pcb{0.04}\% \\ 
        ~ & {\cellcolor[HTML]{F2F2F2}} Fiction \& Literature & {\cellcolor[HTML]{F2F2F2}} \pcb{86.2}\% \\ 
        \midrule
        \multirow{8}{*}{Ratebeer} & {\cellcolor{white}} Strong Ales \& Barley Wines & {\cellcolor{white}} \pcb{8.9}\% \\ 
        ~ & {\cellcolor[HTML]{F2F2F2}} Wheat Beers & {\cellcolor[HTML]{F2F2F2}} \pcb{6.7}\% \\ 
        ~ & {\cellcolor{white}} Dark Beers & {\cellcolor{white}} \pcb{19.7}\% \\ 
        ~ & {\cellcolor[HTML]{F2F2F2}} Lagers & {\cellcolor[HTML]{F2F2F2}} \pcb{16.1}\% \\ 
        ~ & {\cellcolor{white}} Belgian \& Specialty Ales  & {\cellcolor{white}} \pcb{15.8}\% \\ 
        ~ & {\cellcolor[HTML]{F2F2F2}} Pale Ales \& IPAs & {\cellcolor[HTML]{F2F2F2}} \pcb{26.2}\% \\ 
        ~ & {\cellcolor{white}}  Sour \& Fruit Beers  & {\cellcolor{white}} \pcb{5.4}\% \\ 
        ~ & {\cellcolor[HTML]{F2F2F2}} Ciders, Meads \& Sakes & {\cellcolor[HTML]{F2F2F2}} \pcb{1.3}\% \\ 
        \bottomrule
    \end{tabular}
    \end{adjustbox}
\end{table*}


\section{Methods Details}
\label{appx:method}

\subsection{TGNNs}
\label{appx:old-method}

The traditional temporal graph neural network methods are described as follows:

\begin{itemize}[leftmargin=*,itemsep=0.1pt,topsep=0.1pt]
    \item \textbf{JODIE}~\cite{kumar2019predicting} models the future evolution of entity embeddings by forecasting their trajectories over time to predict future interactions and entity states. It employs two interconnected recurrent neural networks to update the dynamic states of entities, along with a projection operation that estimates each entity’s future embedding trajectory.
    \item \textbf{DyRep}~\cite{trivedi2019dyrep} introduces a recurrent framework to continuously update node states after each interaction. Additionally, it features a temporal-attentive aggregation component designed to capture the evolving structural patterns in temporal graphs.
    \item \textbf{TGAT}~\cite{xu2020inductive} generates node representations by aggregating information from each node's temporally relevant neighbors, using a self-attention mechanism. It also integrates a time encoding function to model temporal dependencies within the graph.
    \item \textbf{TGN}~\cite{rossi2020temporal} maintains a dynamic memory for every node, which is updated whenever the node participates in an interaction. This update process involves a message function, a message aggregator, and a memory updater. A dedicated embedding module is then used to produce time-aware node representations based on these evolving memories.
    \item \textbf{CAWN}~\cite{wang2021inductive} first extracts several causal anonymous walks for each node to uncover the causal dynamics of the network and produce relative node identities. These walks are then encoded using recurrent neural networks, and their outputs are aggregated to compute the final node representations.
    \item \textbf{EdgeBank}~\cite{poursafaei2022towards} is a memorization-based method without trainable parameters for transductive temporal link prediction. It records observed interactions in memory and predicts a positive link if the interaction exists in memory, and negative otherwise.
    \item \textbf{TCL}~\cite{wang2021tcl} constructs each node’s interaction sequence by applying a breadth-first search on its temporally dependent sub-graph. It then employs a graph transformer that integrates both topological and temporal cues for learning node representations, further incorporating a cross-attention mechanism to capture interdependencies between interacting nodes.
    \item \textbf{GraphMixer}~\cite{cong2023we} demonstrates that a fixed time encoding function outperforms a trainable one. This fixed function is embedded within a link encoder built on the MLP-Mixer architecture to model temporal interactions. Additionally, a node encoder using neighbor mean-pooling summarizes node features.
    \item \textbf{DyGFormer}~\cite{yu2023towards} generates node representations by leveraging each node’s historical first-hop interactions. It introduces a neighbor co-occurrence encoding mechanism to capture relationships between nodes based on their interaction histories. Additionally, a patching strategy is applied to segment each interaction sequence into smaller patches, enabling the model to effectively utilize longer historical sequences.
    \item \textbf{FreeDyG}~\cite{tian2024freedyg} enhances task performance by adaptively weighting frequency components in the temporal structure's frequency domain. It extracts both high- and low-frequency components of dynamic patterns, then amplifies task-related frequencies and attenuates unrelated signals.
\end{itemize}

\subsection{LLM-as-Enhancer}

This line of work leverages frozen LLMs to enrich semantic representations for TGNNs, while TGNNs remain the final predictive models. 

\begin{itemize}[leftmargin=*,itemsep=0.1pt,topsep=0.1pt]
    \item \textbf{LKD4DyTAG}~\cite{roy2025llm} feeds textual descriptions of temporal graphs into a frozen teacher LLM to extract embeddings, which are then used to guide a student GNN through knowledge distillation, enhancing the GNN with semantic knowledge captured by the LLM.
    \item \textbf{CROSS}~\cite{zhang2025unifying} first generates textual summaries of interaction histories using an LLM, and then employs a semantic-structure co-encoder to help GNNs incorporate the temporal semantic information summarized by the LLM.
\end{itemize}

\subsection{LLM-as-Predictor}
\label{appx:llm-method}

\begin{wrapfigure}{r}{0.35\textwidth}
    \centering
    \includegraphics[width=0.7\linewidth]{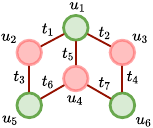}
    \caption{A temporal graph demo.}
    \label{fig:flatten}
\end{wrapfigure}

LLM-as-Predictor approaches directly employ LLMs to solve downstream tasks on temporal graphs. Depending on how LLMs are adapted, this paradigm mainly includes two settings: Supervised Fine-Tuning (SFT) and In-Context Learning (ICL). In both cases, temporal graphs must first be transformed into LLM-understandable sequences, after which temporal structural patterns are learned through either SFT or ICL. 

Existing methods under this paradigm have primarily focused on static graphs and remain under-explored for temporal graphs. Given their strong potential, and to encourage further development of LLM-as-Predictor methods for temporal graph learning, we provide detailed descriptions of the six baselines evaluated in the main paper.

\begin{wraptable}{r}{0.5\textwidth}
    \centering
    \caption{Structural Sequences of Node $u_4$ at Time $t_7$ by ICL and SFT Methods (Based on Figure~\ref{fig:flatten})}
    \label{tab:flatten}
    \resizebox{\linewidth}{!}{
    \begin{tabular}{cc}
    \toprule
    Methods & Sequences of temporal structure \\ 
    \midrule
    ICL & \begin{tabular}[c]{@{}c@{}}$(u_1,u_2,t_1),(u_1,u_3,t_2),(u_2,u_5,t_3),$ \\ $(u_3,u_6,t_4), (u_1,u_4,t_5),(u_4,u_5,t_6)$ \end{tabular} \\ 
    \midrule
    SFT & $(u_1,u_5,u_2,u_3)$ \\ 
    \bottomrule
    \end{tabular}
    }
\end{wraptable}

\textbf{Structural Sequence.} 
\label{param:ss}
Since LLMs can only process token sequences, the structural information in temporal graphs must first be serialized. The ICL-based methods directly represent the temporal structure as a sequence of $(u, v, t)$ triplets sorted in chronological order, while the three SFT-based methods describe the temporal structure of the target node using a multi-hop neighbor sequence obtained through breadth-first search (BFS). Taking the prediction of node $u_4$’s property at time $t_7$ in Figure~\ref{fig:flatten} as an example, the structural sequences generated by ICL and SFT approaches are summarized in Table~\ref{tab:flatten}.

\textbf{Structural Information.} As shown by the serialization strategies in Table~\ref{tab:flatten}, ICL-based methods linearize the temporal structure into a sequence of $(u, v, t)$ triplets, encoding a linear structural representation. This representation is simple and direct, but it requires LLMs to infer temporal structural patterns from the sequential triplets to perform the two temporal graph tasks. In contrast, SFT-based methods represent the temporal structure of the target node through its neighbor sequences, obtained via BFS, which encode explicit hierarchical structural information. This hierarchical information facilitates better structural comprehension by the LLM.

\textbf{Learning and Trainable Parameters.} SFT-based and ICL-based methods employ distinct strategies for LLM training or tuning.
\begin{itemize}[leftmargin=*]
    \item \textit{SFT-based methods fine-tune LLMs to learn structural knowledge for better downstream task performance.} To improve computational efficiency, SFT methods train a projector layer to align temporal structural knowledge with the token embedding space. Specifically, LlaGA-ND and GraphGPT convert the neighbor node sequences into initial text embeddings. LlaGA-HO converts the structural knowledge obtained by aggregating multi-hop neighbor information through message passing from the target node’s embedding. In addition, GraphGPT further fine-tunes the embedding layer of the LLM itself.
    \item \textit{ICL-based methods do not require training or fine-tuning the LLM.} Instead, they leverage the LLM’s pre-trained knowledge and reasoning capabilities, guiding it to understand temporal structural patterns via proposed instructions. To thoroughly explore the LLM’s ability to learn temporal graph structural knowledge through ICL, we adopt the Disentangled Spatial-Temporal Thoughts prompting technique from LLM4DyG~\cite{zhang2024llm4dyg} to TGTalker~\cite{huang2025are}, which encourages the LLM to separately process spatial and temporal information. Specifically, the three variants are: $a$) TGTalker that jointly processes spatial and temporal information; $b$) DST-v1 that processes structural information first, followed by temporal information; and $c$) DST-v2, which processes temporal information first, then structural information. The differences among these three prompting strategies are summarized in Table~\ref{tab:dst}.
\end{itemize}
\begin{table}[t]
    \caption{Differences between three spatial-temporal prompting strategies.}
    \label{tab:dst}
    \begin{tabularx}{0.96\textwidth}{lX}
        \toprule 
          Method           & Prompt  \\
          \midrule 
        TGTalker & Analyze the historical interactions to identify patterns that might indicate a future interaction. Consider both the relationships between users and items and the timing of past interactions within the historical interactions to make your prediction. \\
        \midrule 
        DST-v1 & Think about structure and then time: First analyze the structural information within the historical interactions to understand the relationships between users and items, then consider the temporal patterns to predict the future connection. \\
        \midrule 
        DST-v2 & Think about time and then structure: First analyze the temporal sequence of interactions in the historical data to identify patterns over time, then examine the structural information within the historical interactions to predict the future connection. \\
        \bottomrule 
    \end{tabularx}
\end{table}

\begin{table*}[!t]
    \caption{Prompts of LLM-as-Predictor methods for predicting whether node $u_4$  will form a link with node $u_6$ at time $t_7$, based on the example in Figure~\ref{fig:flatten}}
    \label{tab:flp-prompts}
    \begin{tabularx}{0.96\textwidth}{lX}
        \toprule 
          Method           & Prompt  \\
          \midrule 
        LlaGA-ND & Given two node-centered subgraphs: $\left<u_1,u_5,u_2,u_3\right>$ and $\left<u_3,u_1\right>$, we need to predict whether these two nodes connect with each other. Please tell me whether two center nodes in the subgraphs should connect to each other. \\
        \midrule 
        LlaGA-HO & Given two node-centered subgraphs: $\left<hop_0,hop_1,hop_2\right>$ and $\left<hop_0,hop_1,hop_2\right>$, we need to predict whether these two nodes connect with each other. Please tell me whether two center nodes in the subgraphs should connect to each other. \\
        \midrule 
        GraphGPT           & Given a sequence of graph tokens: $\left<u_1,u_5,u_2,u_3\right>$ that constitute a user-item review subgraph, where the first token represents the central node (the user), and the remaining nodes represent the central node's first- and second-order neighbors. The first-order neighbors are the items that the central node has reviewed and the second-order neighbors are other users who have reviewed the same items. The other sequence of graph tokens: $\left<u_3,u_1\right>$, where the first token corresponds to the center node (the item), and the remaining tokens represent the item's first- and second-order neighbors. The first-order neighbors are the users who have reviewed the item and the second-order neighbors are the items that those users also have reviewed. If the connections between nodes represent the review relationships between users and items, are these two central nodes connected? Give me a direct answer of `yes' or `no'. \\
        \bottomrule 
    \end{tabularx}
\end{table*}

\textbf{Adaptation of SFT-based Methods.} It is important to note that the three SFT methods were originally designed for static graphs. We have adapted them to the temporal graph setting, but due to the increased complexity of temporal graphs, these adaptations cannot fully capture all temporal graph properties. Our adaptations are as follows:

\begin{table*}[t]
    \caption{Prompts of LLM-as-Predictor methods for predicting the label of node $u_4$ at time $t_7$, based on the example in Figure~\ref{fig:flatten}.}
    \label{tab:dnc-prompts}
    \begin{tabularx}{0.96\textwidth}{lX}
        \toprule 
          Method           & Prompt  \\
          \midrule 
        LLaGA-ND & Given a node-centered graph: $\left<u_1,u_5,u_2,u_3\right>$, where the center node represents a user and the other nodes represent the user's historical interactions, including 1\textsuperscript{st}-order neighbors (items that the user has reviewed) and 2\textsuperscript{nd}-order neighbors (other users who have reviewed the same items), classify the center node into one or more of the following 9 interest categories: Side Dishes \& Vegetables, Dietary \& Specialty, Soups \& Stews, Cuisine \& Regional, Desserts \& Sweets, Baking \& Breads, Beverages \& Snacks, Sauces \& Condiments, Main Dishes. \\
        \midrule 
        LLaGA-HO & Given a node-centered graph: $\left<hop_0,hop_1,hop_2\right>$, where the center node represents a user and the other nodes represent the user's historical interactions, including 1\textsuperscript{st}-order neighbors (items that the user has reviewed) and 2\textsuperscript{nd}-order neighbors (other users who have reviewed the same items), classify the center node into one or more of the following 9 interest categories: Side Dishes \& Vegetables, Dietary \& Specialty, Soups \& Stews, Cuisine \& Regional, Desserts \& Sweets, Baking \& Breads, Beverages \& Snacks, Sauces \& Condiments, Main Dishes. \\
        \midrule 
        GraphGPT & Given a user-item review graph: $\left<u_1,u_5,u_2,u_3\right>$ where the 0th node is the target user, and the other nodes are its first- or second-order neighbors. The first-order neighbors are the items that the user has reviewed and the second-order neighbors are other users who have reviewed the same items. Classify the target user's interest into one or more of the following 9 categories: Side Dishes \& Vegetables, Dietary \& Specialty, Soups \& Stews, Cuisine \& Regional, Desserts \& Sweets, Baking \& Breads, Beverages \& Snacks, Sauces \& Condiments, Main Dishes. \\
        \bottomrule 
    \end{tabularx}
\end{table*}

\begin{itemize}[leftmargin=*]
    \item For all methods, the neighbor sequences of the target node are time-sensitive. For example, in Figure~\ref{fig:flatten}, when predicting the property of node $u_4$ at time $t_7$, its first-order neighbors only include $u_1$ and $u_5$, but not $u_6$.
    \item Both LlaGA-HO and GraphGPT require a message-passing process during their Text-Graph Grounding phase, which depends on known graph structures. To accommodate this, we construct a static graph structure by aggregating interactions from the training set, ignoring their timestamps.
    \item Due to computational constraints, we omit the Self-Supervised Instruction Tuning phase in GraphGPT’s training pipeline while retaining the other two phases.
\end{itemize}

Finally, in Table~\ref{tab:flp-prompts} and Table~\ref{tab:dnc-prompts}, we provide detailed prompts for all methods on the two tasks — temporal link prediction and temporal node classification — using the example of predicting whether there is a connection between $u_4$ and $u_6$ at time $t_7$, and predicting the label of $u_4$ at time $t_7$, respectively, as illustrated in Figure~\ref{fig:flatten}.

\section{Experiments Details} 
\label{appx:exp}

\subsection{Setup Details}
\label{appx:setup}

For the TGNN-Predictor methods, we primarily adopt the DyGLib codebase\footnote{\url{https://github.com/yule-BUAA/DyGLib}}. Following their workflow, we first perform a thorough grid search on the hyperparameters of nine traditional methods based on the temporal link prediction task to identify their optimal settings. After training the best-performing models on the temporal link prediction task, we use them as initial checkpoints to continue training for the temporal node classification task.

For the temporal link prediction task, we use supervised binary cross-entropy loss as the objective function. For the temporal node classification task, we treat each label in the FOOD and IMDB multi-label datasets as an independent binary classification problem, applying a weighted binary cross-entropy loss to account for class imbalance. For the other three multi-class datasets, we use a weighted cross-entropy loss as the objective function for the same reason.


For the LLM-Predictor methods, we follow the settings in their original papers as closely as possible, making appropriate adjustments to ensure they could be properly applied to temporal graphs. Specifically, for LlaGA-ND and LlaGA-HO, we consistently set the learning rate to 2e-5 and the batch size to 16 for all models, and trained them for one epoch on each dataset. For LlaGA-HO, since multi-hop embeddings of target nodes require the graph structure, we constructed a static graph by removing timestamp information from edges in the training set, which was then used as the graph structure during embedding computation.

For GraphGPT, due to the complexity of temporal graphs, we retained only two out of its original three pipeline stages: Structural Information Encoding with Text-Graph Grounding and Task-Specific Instruction Fine-tuning. Considering the higher computational cost of Task-Specific Instruction Tuning, we set the batch size to 4 for the Amazon-Kindle dataset and 8 for the other datasets, to fully utilize the 48GB memory of an NVIDIA A6000 GPU. The learning rate was kept at 2e-3, and we set the maximum output length of the LLM to 4096 tokens to accommodate as much structural information as possible. As with LlaGA, this model was also trained for one epoch on each dataset.

\begin{table*}[t]
\caption{Detailed main temporal link prediction results on small-scale datasets. Results are averaged over three independent runs (in \%). }
\centering
\begin{adjustbox}{width=0.98\textwidth}
\begin{tabular}{cccccccccc}
\toprule

\multicolumn{1}{c}{\multirow{2}{*}{Methods}} & \multicolumn{3}{c}{FOOD} & \multicolumn{3}{c}{IMDB} & \multicolumn{3}{c}{Librarything} \\
\multicolumn{1}{c}{} & AP & AUROC & MRR & AP & AUROC & MRR & AP & AUROC & MRR \\
\midrule

JODIE & \ms{72.96}{1.88} & \ms{71.95}{0.69} & \ms{40.09}{0.83} & \ms{60.30}{2.37} & \ms{65.74}{3.20} & \ms{25.57}{1.69} & \ms{75.76}{0.63} & \ms{73.62}{0.75} & \ms{44.14}{0.97} \\ 
DyRep & \ms{72.83}{0.88} & \ms{69.01}{1.53} & \ms{40.55}{0.44} & \ms{64.79}{1.30} & \ms{69.79}{2.03} & \ms{27.05}{0.60} & \ms{73.81}{1.55} & \ms{72.34}{1.13} & \ms{44.13}{0.73} \\ 
TGAT & \ms{54.79}{1.25} & \ms{54.38}{0.67} & \ms{20.14}{1.34} & \ms{54.55}{0.01} & \ms{57.85}{0.05} & \ms{19.23}{0.10} & \ms{64.20}{0.89} & \ms{65.68}{0.96} & \ms{27.74}{1.17} \\ 
TGN & \ms{77.74}{0.05} & \ms{75.74}{0.28} & \ms{44.11}{0.11} & \ms{45.40}{0.89} & \ms{48.79}{1.66} & \ms{10.80}{0.58} & \ms{82.60}{0.22} & \ms{82.02}{0.09} & \ms{52.29}{0.46} \\ 
CAWN & \ms{51.71}{0.38} & \ms{53.96}{0.23} & \ms{17.43}{0.38} & \ms{54.00}{1.17} & \ms{55.60}{2.75} & \ms{19.52}{0.49} & \ms{62.01}{0.21} & \ms{63.21}{0.07} & \ms{26.70}{0.08} \\ 
TCL & \ms{49.12}{0.79} & \ms{50.98}{1.23} & \ms{15.43}{0.55} & \ms{55.78}{0.49} & \ms{58.83}{0.44} & \ms{20.18}{0.21} & \ms{54.51}{4.02} & \ms{56.63}{2.91} & \ms{18.88}{4.14} \\ 
GraphMixer & \ms{66.98}{1.71} & \ms{65.78}{0.93} & \ms{31.62}{1.98} & \ms{57.01}{0.03} & \ms{61.60}{0.09} & \ms{21.20}{0.05} & \ms{62.13}{1.45} & \ms{63.16}{2.44} & \ms{26.52}{1.26} \\ 
DyGFormer & \ms{59.48}{0.24} & \ms{57.26}{0.52} & \ms{24.51}{0.13} & \ms{55.32}{0.11} & \ms{58.42}{0.10} & \ms{20.24}{0.02} & \ms{54.53}{2.16} & \ms{54.35}{3.14} & \ms{19.75}{2.34} \\ 
FreeDyG & \ms{72.40}{0.34} & \ms{69.60}{0.37} & \ms{38.56}{0.52} & \ms{57.53}{0.89} & \ms{62.16}{0.93} & \ms{21.78}{0.80} & \ms{66.08}{0.26} & \ms{65.13}{1.70} & \ms{30.90}{0.13} \\ 
LKD4DyTAG & \ms{66.63}{0.95} & \ms{63.62}{0.95} & \ms{31.30}{1.05} & \ms{56.04}{0.92} & \ms{59.13}{0.53} & \ms{20.50}{0.76} & \ms{59.94}{0.49} & \ms{62.17}{0.34} & \ms{23.37}{0.35} \\ 
\midrule
CROSS & \ms{58.15}{0.13} & \ms{56.48}{0.24} & \ms{23.17}{0.23} & \ms{55.98}{0.05} & \ms{59.14}{0.12} & \ms{20.62}{0.06} & \ms{59.04}{0.83} & \ms{60.25}{0.94} & \ms{24.16}{0.53} \\ 
TGTalker & \ms{51.04}{0.02} & \ms{52.02}{0.03} & \ms{16.24}{0.15} & \ms{50.22}{0.04} & \ms{50.43}{0.07} & \ms{15.12}{0.10} & \ms{50.63}{ 0.01} & \ms{51.24}{0.04} & \ms{15.54}{0.12} \\ 
DST-v1 & \ms{51.31}{0.01} & \ms{52.53}{0.03} & \ms{16.58}{0.18} & \ms{50.26}{0.10} & \ms{50.52}{0.20} & \ms{15.18}{0.22} & \ms{50.71}{0.02} & \ms{51.39}{0.06} & \ms{15.68}{0.15} \\ 
DST-v2 & \ms{51.33}{0.00} & \ms{52.57}{0.01} & \ms{16.60}{0.15} & \ms{50.35}{0.10} & \ms{50.70}{0.19} & \ms{15.25}{0.18} & \ms{50.61}{0.01} & \ms{51.19}{0.02} & \ms{15.51}{0.08} \\ 
Llaga-ND & \ms{53.10}{0.36} & \ms{55.52}{0.57} & \ms{18.85}{0.55} & \ms{52.01}{0.75} & \ms{53.55}{1.19} & \ms{17.50}{0.85} & \ms{49.98}{0.04} & \ms{49.95}{0.07} & \ms{14.85}{0.10} \\ 
Llaga-HO & \ms{61.95}{1.32} & \ms{65.67}{1.47} & \ms{29.50}{1.85} & \ms{54.37}{0.82} & \ms{57.34}{1.06} & \ms{20.15}{1.25} & \ms{49.97}{0.04} & \ms{49.95}{0.07} & \ms{14.83}{0.11} \\ 
GraphGPT & \ms{55.09}{0.43} & \ms{57.95}{1.02} & \ms{25.05}{0.85} & \ms{51.80}{0.18} & \ms{53.19}{0.29} & \ms{17.20}{0.45} & \ms{49.96}{0.04} & \ms{49.89}{0.11} & \ms{14.81}{0.18} \\ 
\bottomrule
\end{tabular}
\end{adjustbox}
\label{tab:flp-detail-1}
\end{table*}

\begin{table*}[t]
\caption{Detailed main temporal link prediction results on large-scale datasets. Results are averaged over three independent runs (in \%). }
\centering
\begin{adjustbox}{width=0.98\textwidth}
\begin{tabular}{cccccccccc}
\toprule
\multicolumn{1}{c}{\multirow{2}{*}{Methods}} & \multicolumn{3}{c}{Beeradvocate} & \multicolumn{3}{c}{Amazon-Kindle} & \multicolumn{3}{c}{Ratebeer} \\
\multicolumn{1}{c}{} & AP & AUROC & MRR & AP & AUROC & MRR & AP & AUROC & MRR \\
\midrule
JODIE & \ms{79.75}{11.63} & \ms{80.93}{9.53} & \ms{57.49}{16.89} & \ms{82.81}{2.00} & \ms{81.86}{1.69} & \ms{60.05}{3.18} & \ms{92.62}{0.18} & \ms{91.56}{0.36} & \ms{74.20}{0.10} \\ 
DyRep & \ms{89.61}{0.20} & \ms{87.01}{0.40} & \ms{72.50}{0.16} & \ms{78.35}{2.63} & \ms{76.63}{2.69} & \ms{56.99}{3.23} & \ms{85.87}{0.33} & \ms{83.84}{1.18} & \ms{66.11}{0.71} \\ 
TGAT & \ms{59.37}{3.59} & \ms{66.58}{2.13} & \ms{27.90}{4.64} & \ms{79.00}{1.97} & \ms{81.12}{1.32} & \ms{44.37}{2.71} & \ms{58.93}{4.20} & \ms{64.38}{2.29} & \ms{28.63}{3.66} \\ 
TGN & \ms{83.46}{1.68} & \ms{83.25}{1.06} & \ms{59.35}{2.48} & \ms{71.77}{2.90} & \ms{79.92}{1.49} & \ms{39.67}{3.72} & \ms{91.55}{0.24} & \ms{91.41}{0.24} & \ms{72.00}{0.31} \\ 
CAWN & \ms{66.15}{0.85} & \ms{71.29}{0.05} & \ms{35.12}{0.20} & \ms{79.34}{1.50} & \ms{80.79}{1.22} & \ms{44.92}{2.04} & \ms{59.94}{2.40} & \ms{65.78}{0.67} & \ms{29.86}{1.58} \\ 
TCL & \ms{55.40}{0.18} & \ms{64.62}{0.29} & \ms{22.37}{1.01} & \ms{79.68}{1.51} & \ms{80.96}{0.95} & \ms{44.74}{2.60} & \ms{57.61}{3.77} & \ms{62.63}{2.39} & \ms{24.94}{4.85} \\ 
GraphMixer & \ms{86.02}{0.24} & \ms{85.57}{0.29} & \ms{60.51}{0.84} & \ms{81.53}{0.28} & \ms{81.87}{0.09} & \ms{48.81}{0.44} & \ms{78.53}{4.22} & \ms{78.05}{4.06} & \ms{48.52}{7.06} \\ 
DyGFormer & \ms{50.53}{0.08} & \ms{56.80}{0.05} & \ms{16.34}{0.71} & \ms{72.78}{0.32} & \ms{74.71}{0.09} & \ms{36.61}{0.22} & \ms{53.28}{0.91} & \ms{57.65}{0.73} & \ms{23.09}{1.33} \\ 
FreeDyG & \ms{82.93}{6.75} & \ms{85.18}{4.07} & \ms{58.86}{8.56} & \ms{67.26}{3.83} & \ms{71.25}{1.99} & \ms{32.80}{4.85} & \ms{69.49}{3.30} & \ms{74.16}{1.86} & \ms{40.10}{3.65} \\ 
LKD4DyTAG & \ms{79.42}{1.15} & \ms{79.52}{0.91} & \ms{47.51}{1.41} & \ms{80.64}{0.46} & \ms{81.64}{0.42} & \ms{46.44}{0.59} & \ms{69.51}{0.03} & \ms{70.85}{0.05} & \ms{37.59}{0.07} \\ 
CROSS & \ms{51.36}{1.03} & \ms{57.52}{1.01} & \ms{17.89}{2.25} & \ms{87.25}{0.13} & \ms{88.31}{0.12} & \ms{56.99}{0.38} & \ms{52.57}{0.09} & \ms{57.19}{0.21} & \ms{21.32}{0.98} \\ 
\midrule
TGTalker & \ms{51.18}{0.03} & \ms{51.36}{0.02} & \ms{16.05}{0.08} & \ms{52.54}{0.09} & \ms{54.79}{0.11} & \ms{18.21}{0.35} & \ms{51.11}{0.07} & \ms{52.15}{0.06} & \ms{16.10}{0.14} \\ 
DST-v1 & \ms{51.34}{0.04} & \ms{51.47}{0.05} & \ms{16.18}{0.10} & \ms{52.66}{0.10} & \ms{54.85}{0.09} & \ms{18.32}{0.41} & \ms{51.23}{0.04} & \ms{51.26}{0.03} & \ms{16.15}{0.09} \\ 
DST-v2 & \ms{51.21}{0.01} & \ms{51.53}{0.01} & \ms{16.12}{0.05} & \ms{52.62}{0.08} & \ms{54.81}{0.08} & \ms{18.28}{0.38} & \ms{51.05}{0.01} & \ms{51.07}{0.02} & \ms{15.98}{0.05} \\ 
Llaga-ND & \ms{49.97}{0.03} & \ms{49.94}{0.06} & \ms{14.82}{0.12} & \ms{68.62}{3.37} & \ms{72.85}{3.58} & \ms{36.45}{4.12} & \ms{50.00}{0.02} & \ms{49.99}{0.05} & \ms{14.90}{0.08} \\ 
Llaga-HO & \ms{50.00}{0.05} & \ms{50.00}{0.10} & \ms{14.91}{0.15} & \ms{72.64}{1.11} & \ms{76.63}{1.54} & \ms{42.10}{2.05} & \ms{50.00}{0.00} & \ms{50.00}{0.00} & \ms{15.07}{0.10} \\ 
GraphGPT & \ms{56.80}{6.80} & \ms{58.35}{8.35} & \ms{22.15}{7.50} & \ms{50.72}{0.72} & \ms{51.17}{1.17} & \ms{15.75}{1.05} & \ms{56.03}{6.03} & \ms{59.29}{9.29} & \ms{21.80}{8.15} \\    	 	

\bottomrule
\end{tabular}
\end{adjustbox}
\label{tab:flp-detail-2}
\end{table*}

\begin{table*}[!h]
\caption{Temporal link prediction results on small-scale datasets under inductive settings. Results are averaged over three independent runs (in \%). }
\centering
\begin{adjustbox}{width=0.98\textwidth}
\begin{tabular}{cccccccccc}
\toprule
\multicolumn{1}{c}{\multirow{2}{*}{Methods}} & \multicolumn{3}{c}{FOOD} & \multicolumn{3}{c}{IMDB} & \multicolumn{3}{c}{Librarything} \\
\multicolumn{1}{c}{} & AP & AUROC & MRR & AP & AUROC & MRR & AP & AUROC & MRR \\
\midrule

JODIE & \ms{76.02}{0.07} & \ms{74.28}{0.06} & \ms{41.71}{0.32} & \ms{88.39}{0.21} & \ms{87.47}{0.35} & \ms{61.51}{0.31} & \ms{84.46}{0.02} & \ms{84.47}{0.08} & \ms{57.68}{0.63} \\ 
DyRep & \ms{73.08}{0.91} & \ms{71.48}{0.85} & \ms{37.38}{1.46} & \ms{86.79}{0.49} & \ms{85.72}{0.60} & \ms{59.30}{0.41} & \ms{82.45}{0.16} & \ms{82.96}{0.21} & \ms{54.40}{0.21} \\ 
TGAT & \ms{54.75}{0.65} & \ms{54.30}{0.23} & \ms{20.12}{0.67} & \ms{74.53}{0.21} & \ms{73.75}{0.30} & \ms{39.85}{0.06} & \ms{71.31}{0.95} & \ms{71.94}{0.93} & \ms{33.97}{1.28} \\ 
TGN & \ms{73.77}{0.44} & \ms{72.14}{0.46} & \ms{38.43}{0.30} & \ms{79.66}{1.92} & \ms{79.23}{2.39} & \ms{47.26}{1.20} & \ms{83.38}{0.03} & \ms{84.35}{0.20} & \ms{54.87}{0.16} \\ 
CAWN & \ms{53.06}{0.41} & \ms{54.33}{0.50} & \ms{18.74}{0.35} & \ms{63.64}{0.95} & \ms{66.12}{0.27} & \ms{27.52}{1.43} & \ms{68.74}{1.25} & \ms{69.21}{1.06} & \ms{31.96}{1.66} \\ 
TCL & \ms{50.40}{0.75} & \ms{50.71}{1.14} & \ms{16.85}{0.42} & \ms{67.39}{3.46} & \ms{68.20}{1.39} & \ms{31.95}{4.87} & \ms{64.02}{2.34} & \ms{65.14}{1.62} & \ms{26.35}{2.96} \\ 
GraphMixer & \ms{68.41}{1.54} & \ms{67.19}{1.10} & \ms{33.08}{1.68} & \ms{86.67}{0.03} & \ms{84.91}{0.02} & \ms{60.07}{0.03} & \ms{77.25}{1.73} & \ms{76.95}{2.01} & \ms{43.78}{2.03} \\ 
DyGFormer & \ms{58.00}{0.21} & \ms{56.37}{0.44} & \ms{23.03}{0.10} & \ms{62.33}{0.38} & \ms{65.88}{0.09} & \ms{25.71}{0.48} & \ms{61.16}{3.16} & \ms{60.93}{3.65} & \ms{24.63}{2.80} \\ 
FreeDyG & \ms{72.65}{0.18} & \ms{70.38}{0.13} & \ms{38.47}{0.35} & \ms{86.90}{0.08} & \ms{85.15}{0.10} & \ms{60.26}{0.07} & \ms{79.81}{0.71} & \ms{78.54}{1.32} & \ms{47.81}{0.98} \\ 
LKD4DyTAG & \ms{62.13}{0.67} & \ms{59.74}{0.65} & \ms{26.74}{0.61} & \ms{61.81}{0.76} & \ms{65.71}{0.22} & \ms{25.20}{0.80} & \ms{67.44}{0.19} & \ms{68.06}{0.24} & \ms{29.98}{0.03} \\ 
CROSS & \ms{56.77}{0.44} & \ms{55.39}{0.28} & \ms{21.95}{0.35} & \ms{62.27}{1.18} & \ms{66.97}{0.47} & \ms{25.16}{1.39} & \ms{65.79}{0.51} & \ms{66.24}{0.53} & \ms{28.82}{0.41} \\ 
\midrule
TGTalker & \ms{51.04}{0.02} & \ms{52.02}{0.03} & \ms{16.20}{0.12} & \ms{50.22}{0.04} & \ms{50.43}{0.07} & \ms{15.15}{0.11} & \ms{50.63}{ 0.01} & \ms{51.24}{0.04} & \ms{15.50}{0.08} \\ 
DST-v1 & \ms{51.31}{0.01} & \ms{52.53}{0.03} & \ms{16.55}{0.15} & \ms{50.26}{0.10} & \ms{50.52}{0.20} & \ms{15.20}{0.20} & \ms{50.71}{0.02} & \ms{51.39}{0.06} & \ms{15.65}{0.12} \\ 
DST-v2 & \ms{51.33}{0.00} & \ms{52.57}{0.01} & \ms{16.58}{0.10} & \ms{50.35}{0.10} & \ms{50.70}{0.19} & \ms{15.28}{0.15} & \ms{50.61}{0.01} & \ms{51.19}{0.02} & \ms{15.48}{0.06} \\ 
Llaga-ND & \ms{53.10}{0.36} & \ms{55.52}{0.57} & \ms{18.80}{0.50} & \ms{52.01}{0.75} & \ms{53.55}{1.19} & \ms{17.45}{0.80} & \ms{49.98}{0.04} & \ms{49.95}{0.07} & \ms{14.85}{0.12} \\ 
Llaga-HO & \ms{61.95}{1.32} & \ms{65.67}{1.47} & \ms{29.45}{1.80} & \ms{54.37}{0.82} & \ms{57.34}{1.06} & \ms{20.10}{1.20} & \ms{49.97}{0.04} & \ms{49.95}{0.07} & \ms{14.83}{0.10} \\ 
GraphGPT & \ms{55.09}{0.43} & \ms{57.95}{1.02} & \ms{25.00}{0.80} & \ms{51.80}{0.18} & \ms{53.19}{0.29} & \ms{17.15}{0.40} & \ms{49.96}{0.04} & \ms{49.89}{0.11} & \ms{14.80}{0.15} \\ 	

\bottomrule
\end{tabular}
\end{adjustbox}
\label{tab:lp-ind-small}
\end{table*}

\begin{table*}[!t]
\caption{Temporal link prediction results on large-scale datasets under inductive settings. Results are averaged over three independent runs (in \%). }
\centering
\begin{adjustbox}{width=0.98\textwidth}
\begin{tabular}{cccccccccc}
\toprule
\multicolumn{1}{c}{\multirow{2}{*}{Methods}} & \multicolumn{3}{c}{Beeradvocate} & \multicolumn{3}{c}{Amazon-Kindle} & \multicolumn{3}{c}{Ratebeer} \\
\multicolumn{1}{c}{} & AP & AUROC & MRR & AP & AUROC & MRR & AP & AUROC & MRR \\
\midrule

JODIE & \ms{90.84}{0.51} & \ms{89.99}{0.68} & \ms{68.66}{0.96} & \ms{93.78}{0.38} & \ms{93.01}{0.37} & \ms{78.89}{0.98} & \ms{90.38}{0.70} & \ms{89.81}{0.77} & \ms{67.92}{1.10} \\ 
DyRep & \ms{85.81}{2.03} & \ms{84.59}{1.78} & \ms{62.14}{3.80} & \ms{91.25}{0.43} & \ms{90.17}{0.46} & \ms{73.40}{0.85} & \ms{83.59}{0.11} & \ms{81.82}{0.08} & \ms{57.22}{0.03} \\ 
TGAT & \ms{59.67}{3.80} & \ms{62.48}{2.60} & \ms{29.21}{4.06} & \ms{82.51}{1.03} & \ms{83.97}{0.58} & \ms{49.22}{1.64} & \ms{62.19}{3.89} & \ms{64.92}{2.37} & \ms{31.86}{3.16} \\ 
TGN & \ms{79.71}{8.75} & \ms{80.85}{7.37} & \ms{52.83}{13.11} & \ms{94.14}{0.51} & \ms{94.43}{0.36} & \ms{77.84}{0.14} & \ms{86.56}{0.87} & \ms{86.10}{1.15} & \ms{61.47}{1.89} \\ 
CAWN & \ms{66.89}{0.95} & \ms{68.29}{0.09} & \ms{36.61}{0.28} & \ms{80.89}{0.11} & \ms{81.90}{0.18} & \ms{46.82}{0.33} & \ms{63.92}{2.23} & \ms{66.57}{0.79} & \ms{33.90}{1.23} \\ 
TCL & \ms{54.64}{0.16} & \ms{59.35}{0.57} & \ms{24.08}{1.07} & \ms{81.60}{0.22} & \ms{82.53}{0.13} & \ms{47.51}{0.67} & \ms{61.99}{4.64} & \ms{63.42}{3.14} & \ms{30.17}{5.63} \\ 
GraphMixer & \ms{87.00}{0.19} & \ms{85.52}{0.26} & \ms{62.75}{0.59} & \ms{93.33}{0.04} & \ms{92.71}{0.03} & \ms{74.54}{0.11} & \ms{82.19}{3.88} & \ms{80.38}{4.26} & \ms{54.45}{6.40} \\ 
DyGFormer & \ms{49.72}{0.31} & \ms{52.27}{0.24} & \ms{17.36}{0.81} & \ms{74.79}{0.40} & \ms{76.32}{0.17} & \ms{38.63}{0.31} & \ms{57.53}{0.92} & \ms{59.29}{0.80} & \ms{26.90}{1.26} \\ 
FreeDyG & \ms{84.22}{5.36} & \ms{84.83}{3.81} & \ms{60.85}{5.85} & \ms{87.64}{1.74} & \ms{86.77}{1.48} & \ms{64.46}{4.09} & \ms{76.08}{2.92} & \ms{77.48}{2.02} & \ms{48.59}{3.36} \\ 
LKD4DyTAG & \ms{76.66}{0.82} & \ms{75.61}{0.82} & \ms{43.50}{0.81} & \ms{80.48}{0.07} & \ms{81.69}{0.12} & \ms{46.02}{0.17} & \ms{71.56}{0.16} & \ms{70.95}{0.14} & \ms{39.32}{0.22} \\ 
CROSS & \ms{50.99}{1.61} & \ms{53.34}{1.42} & \ms{19.13}{2.48} & \ms{87.47}{0.22} & \ms{88.73}{0.21} & \ms{57.42}{0.36} & \ms{56.94}{0.11} & \ms{59.06}{0.06} & \ms{25.55}{0.74} \\ 
\midrule
TGTalker & \ms{51.18}{0.03} & \ms{51.36}{0.02} & \ms{16.00}{0.09} & \ms{52.54}{0.09} & \ms{54.79}{0.11} & \ms{18.15}{0.30} & \ms{52.41}{0.13} & \ms{54.09}{3.92} & \ms{17.80}{2.50} \\ 
DST-v1 & \ms{51.34}{0.04} & \ms{51.47}{0.05} & \ms{16.15}{0.11} & \ms{52.66}{0.10} & \ms{54.85}{0.09} & \ms{18.25}{0.38} & \ms{52.96}{0.94} & \ms{55.60}{1.22} & \ms{18.65}{1.05} \\ 
DST-v2 & \ms{51.21}{0.01} & \ms{51.53}{0.01} & \ms{16.10}{0.04} & \ms{52.62}{0.08} & \ms{54.81}{0.08} & \ms{18.20}{0.35} & \ms{51.72}{1.08} & \ms{54.15}{2.89} & \ms{17.15}{1.85} \\ 
Llaga-ND & \ms{49.97}{0.03} & \ms{49.94}{0.06} & \ms{14.80}{0.15} & \ms{68.62}{3.37} & \ms{72.85}{3.58} & \ms{36.40}{4.10} & \ms{50.00}{0.02} & \ms{49.99}{0.05} & \ms{14.90}{0.05} \\ 
Llaga-HO & \ms{50.00}{0.05} & \ms{50.00}{0.10} & \ms{14.90}{0.12} & \ms{72.64}{1.11} & \ms{76.63}{1.54} & \ms{42.05}{2.00} & \ms{50.00}{0.00} & \ms{50.00}{0.00} & \ms{14.90}{0.00} \\ 
GraphGPT & \ms{56.80}{6.80} & \ms{58.35}{8.35} & \ms{22.10}{7.45} & \ms{50.72}{0.72} & \ms{51.17}{1.17} & \ms{15.70}{1.00} & \ms{56.03}{6.03} & \ms{59.29}{9.29} & \ms{27.75}{8.10} \\  

\bottomrule
\end{tabular}
\end{adjustbox}
\label{tab:lp-ind-large}
\end{table*}

\begin{table*}[t]
\caption{Temporal link prediction results on small-scale datasets under transductive settings with historical negative sampling. Results are averaged over three independent runs (in \%). }
\centering
\begin{adjustbox}{width=0.98\textwidth}
\begin{tabular}{cccccccccc}
\toprule
\multicolumn{1}{c}{\multirow{2}{*}{Methods}} & \multicolumn{3}{c}{FOOD} & \multicolumn{3}{c}{IMDB} & \multicolumn{3}{c}{Librarything} \\
\multicolumn{1}{c}{} & AP & AUROC & MRR & AP & AUROC & MRR & AP & AUROC & MRR \\
\midrule

JODIE & \ms{55.00}{0.11} & \ms{55.13}{0.30} & \ms{20.64}{0.17} & \ms{66.09}{8.32} & \ms{72.60}{7.93} & \ms{30.45}{9.29} & \ms{53.86}{1.56} & \ms{53.43}{1.58} & \ms{19.97}{1.10} \\ 
DyRep & \ms{55.13}{0.49} & \ms{53.93}{0.49} & \ms{21.10}{0.40} & \ms{57.57}{3.78} & \ms{57.83}{4.89} & \ms{23.09}{2.91} & \ms{56.62}{1.13} & \ms{56.83}{0.87} & \ms{21.91}{0.93} \\ 
TGAT & \ms{51.38}{0.05} & \ms{50.69}{0.09} & \ms{17.92}{0.04} & \ms{46.59}{0.37} & \ms{45.36}{1.13} & \ms{14.18}{0.13} & \ms{45.51}{0.06} & \ms{45.35}{0.22} & \ms{13.11}{0.16} \\ 
TGN & \ms{55.73}{0.21} & \ms{57.26}{0.23} & \ms{20.58}{0.16} & \ms{40.40}{0.71} & \ms{37.98}{1.74} & \ms{8.39}{0.32} & \ms{53.66}{0.36} & \ms{54.44}{0.17} & \ms{19.25}{0.30} \\ 
CAWN & \ms{52.26}{0.01} & \ms{52.06}{0.04} & \ms{18.50}{0.05} & \ms{45.85}{0.25} & \ms{43.61}{0.47} & \ms{14.19}{0.07} & \ms{48.06}{0.16} & \ms{48.74}{0.10} & \ms{14.98}{0.10} \\ 
TCL & \ms{50.49}{0.04} & \ms{49.83}{0.12} & \ms{17.29}{0.14} & \ms{46.07}{1.06} & \ms{44.17}{0.93} & \ms{14.32}{0.86} & \ms{46.04}{0.32} & \ms{46.53}{0.54} & \ms{13.39}{0.22} \\ 
GraphMixer & \ms{58.75}{1.17} & \ms{57.56}{1.90} & \ms{23.79}{0.76} & \ms{51.95}{0.26} & \ms{57.20}{0.77} & \ms{16.65}{0.03} & \ms{51.01}{0.28} & \ms{49.59}{0.55} & \ms{17.90}{0.41} \\ 
DyGFormer & \ms{50.85}{0.66} & \ms{49.79}{1.16} & \ms{17.57}{0.36} & \ms{46.75}{0.28} & \ms{44.53}{0.50} & \ms{14.80}{0.25} & \ms{49.37}{1.25} & \ms{50.20}{0.52} & \ms{16.00}{1.09} \\ 
FreeDyG & \ms{57.66}{0.31} & \ms{55.46}{0.59} & \ms{23.30}{0.22} & \ms{55.06}{0.83} & \ms{59.51}{1.48} & \ms{19.67}{0.30} & \ms{51.03}{0.45} & \ms{49.23}{0.00} & \ms{18.06}{0.38} \\ 
LKD4DyTAG & \ms{52.30}{0.00} & \ms{51.96}{0.08} & \ms{18.62}{0.02} & \ms{45.77}{0.60} & \ms{42.98}{0.36} & \ms{14.12}{0.46} & \ms{46.09}{0.14} & \ms{46.16}{0.21} & \ms{13.71}{0.12} \\ 
CROSS & \ms{50.86}{0.28} & \ms{49.75}{0.11} & \ms{17.66}{0.21} & \ms{47.71}{0.04} & \ms{45.20}{0.02} & \ms{15.63}{0.07} & \ms{46.78}{0.05} & \ms{47.45}{0.08} & \ms{14.03}{0.06} \\  

\bottomrule
\end{tabular}
\end{adjustbox}
\label{tab:lp-trand-small-his}
\end{table*}

\begin{table*}[t]
\caption{Temporal link prediction results on small-scale datasets under inductive settings with historical negative sampling. Results are averaged over three independent runs (in \%). }
\centering
\begin{adjustbox}{width=0.98\textwidth}
\begin{tabular}{cccccccccc}
\toprule
\multicolumn{1}{c}{\multirow{2}{*}{Methods}} & \multicolumn{3}{c}{FOOD} & \multicolumn{3}{c}{IMDB} & \multicolumn{3}{c}{Librarything} \\
\multicolumn{1}{c}{} & AP & AUROC & MRR & AP & AUROC & MRR & AP & AUROC & MRR \\
\midrule

JODIE & \ms{50.08}{0.39} & \ms{46.65}{0.10} & \ms{17.65}{0.47} & \ms{74.19}{0.85} & \ms{71.07}{1.83} & \ms{41.40}{0.35} & \ms{52.27}{0.06} & \ms{47.60}{0.69} & \ms{19.80}{0.09} \\ 
DyRep & \ms{48.32}{0.29} & \ms{45.69}{0.06} & \ms{16.11}{0.29} & \ms{69.73}{0.52} & \ms{65.67}{1.27} & \ms{36.57}{0.31} & \ms{50.86}{0.17} & \ms{46.70}{0.74} & \ms{18.40}{0.04} \\ 
TGAT & \ms{51.01}{0.25} & \ms{50.25}{0.03} & \ms{17.66}{0.30} & \ms{64.16}{0.67} & \ms{60.85}{0.71} & \ms{30.34}{0.56} & \ms{47.20}{0.22} & \ms{45.20}{0.09} & \ms{15.05}{0.15} \\ 
TGN & \ms{49.07}{0.34} & \ms{46.38}{0.32} & \ms{16.72}{0.23} & \ms{69.85}{2.88} & \ms{64.18}{1.71} & \ms{37.97}{4.38} & \ms{46.31}{0.19} & \ms{43.16}{0.34} & \ms{14.64}{0.16} \\ 
CAWN & \ms{54.97}{0.22} & \ms{53.61}{0.08} & \ms{20.31}{0.22} & \ms{55.21}{1.39} & \ms{54.88}{1.44} & \ms{21.38}{1.18} & \ms{50.89}{1.13} & \ms{50.06}{1.36} & \ms{17.55}{0.81} \\ 
TCL & \ms{53.10}{0.17} & \ms{49.76}{0.48} & \ms{19.31}{0.30} & \ms{57.06}{4.53} & \ms{54.56}{2.23} & \ms{23.76}{4.87} & \ms{47.35}{0.47} & \ms{44.99}{0.48} & \ms{15.14}{0.41} \\ 
GraphMixer & \ms{56.46}{1.19} & \ms{53.37}{1.60} & \ms{22.80}{0.94} & \ms{75.58}{0.02} & \ms{71.56}{0.10} & \ms{44.36}{0.21} & \ms{61.46}{2.92} & \ms{59.97}{2.19} & \ms{26.70}{2.84} \\ 
DyGFormer & \ms{48.37}{1.76} & \ms{46.33}{3.88} & \ms{15.92}{0.80} & \ms{53.50}{0.41} & \ms{53.75}{0.04} & \ms{19.53}{0.49} & \ms{53.57}{0.32} & \ms{56.43}{0.14} & \ms{18.51}{0.29} \\ 
FreeDyG & \ms{54.50}{0.49} & \ms{50.54}{1.00} & \ms{21.36}{0.24} & \ms{76.96}{0.31} & \ms{73.35}{0.45} & \ms{46.00}{0.55} & \ms{58.12}{1.22} & \ms{56.72}{1.11} & \ms{23.62}{1.13} \\ 
LKD4DyTAG & \ms{50.34}{0.08} & \ms{49.50}{0.13} & \ms{17.12}{0.01} & \ms{51.36}{0.19} & \ms{51.15}{0.01} & \ms{17.98}{0.19} & \ms{48.17}{0.04} & \ms{46.87}{0.00} & \ms{15.57}{0.01} \\ 
CROSS & \ms{49.78}{0.42} & \ms{48.83}{0.34} & \ms{16.78}{0.33} & \ms{55.17}{1.15} & \ms{56.38}{0.67} & \ms{20.51}{1.17} & \ms{50.15}{0.24} & \ms{50.27}{0.43} & \ms{16.61}{0.14} \\ 

\bottomrule
\end{tabular}
\end{adjustbox}
\label{tab:lp-ind-small-his}
\end{table*}

\begin{table*}[!h]
\caption{Temporal link prediction results on large-scale datasets under transductive settings with historical negative sampling. Results are averaged over three independent runs (in \%). }
\centering
\begin{adjustbox}{width=0.98\textwidth}
\begin{tabular}{cccccccccc}
\toprule
\multicolumn{1}{c}{\multirow{2}{*}{Methods}} & \multicolumn{3}{c}{Beeradvocate} & \multicolumn{3}{c}{Amazon-Kindle} & \multicolumn{3}{c}{Ratebeer} \\
\multicolumn{1}{c}{} & AP & AUROC & MRR & AP & AUROC & MRR & AP & AUROC & MRR \\
\midrule

JODIE & \ms{57.46}{3.08} & \ms{62.36}{0.84} & \ms{21.12}{3.28} & \ms{68.84}{1.66} & \ms{68.96}{1.42} & \ms{32.45}{1.80} & \ms{64.03}{0.51} & \ms{67.15}{0.24} & \ms{26.89}{0.48} \\ 
DyRep & \ms{68.04}{2.86} & \ms{71.01}{2.67} & \ms{32.16}{2.35} & \ms{65.26}{0.94} & \ms{64.98}{1.07} & \ms{29.23}{0.81} & \ms{69.96}{2.88} & \ms{72.18}{3.36} & \ms{33.95}{2.39} \\ 
TGAT & \ms{52.55}{0.13} & \ms{52.82}{0.14} & \ms{18.40}{0.09} & \ms{53.77}{0.52} & \ms{54.89}{0.54} & \ms{19.24}{0.37} & \ms{50.26}{0.08} & \ms{49.82}{0.23} & \ms{17.01}{0.00} \\ 
TGN & \ms{53.54}{2.03} & \ms{57.49}{0.89} & \ms{18.18}{1.89} & \ms{44.79}{0.59} & \ms{46.94}{0.86} & \ms{11.13}{0.64} & \ms{57.87}{3.88} & \ms{62.78}{2.82} & \ms{21.51}{3.49} \\ 
CAWN & \ms{52.34}{0.05} & \ms{53.17}{0.01} & \ms{18.26}{0.04} & \ms{54.31}{0.90} & \ms{54.83}{1.17} & \ms{19.78}{0.63} & \ms{51.02}{0.18} & \ms{51.85}{0.10} & \ms{17.30}{0.19} \\ 
TCL & \ms{52.68}{0.17} & \ms{52.85}{0.18} & \ms{18.57}{0.13} & \ms{55.02}{0.81} & \ms{54.65}{0.78} & \ms{20.49}{0.61} & \ms{53.63}{1.84} & \ms{52.81}{1.25} & \ms{19.66}{1.55} \\ 
GraphMixer & \ms{54.82}{1.62} & \ms{57.02}{1.22} & \ms{19.70}{1.33} & \ms{64.62}{0.03} & \ms{65.97}{0.01} & \ms{28.16}{0.06} & \ms{54.79}{0.71} & \ms{55.97}{1.02} & \ms{19.96}{0.48} \\ 
DyGFormer & \ms{52.47}{0.24} & \ms{52.58}{0.14} & \ms{18.45}{0.20} & \ms{53.75}{0.64} & \ms{54.18}{0.53} & \ms{19.29}{0.50} & \ms{53.67}{0.18} & \ms{53.52}{0.33} & \ms{19.43}{0.14} \\ 
FreeDyG & \ms{50.56}{1.71} & \ms{54.11}{0.91} & \ms{15.86}{1.66} & \ms{65.16}{0.09} & \ms{65.98}{0.29} & \ms{28.79}{0.18} & \ms{54.35}{0.29} & \ms{57.11}{0.04} & \ms{19.33}{0.29} \\ 
LKD4DyTAG & \ms{51.97}{0.10} & \ms{52.35}{0.13} & \ms{18.01}{0.05} & \ms{54.95}{0.50} & \ms{55.12}{0.63} & \ms{20.29}{0.38} & \ms{49.85}{0.16} & \ms{49.30}{0.28} & \ms{16.74}{0.11} \\ 
CROSS & \ms{52.14}{0.34} & \ms{52.16}{0.30} & \ms{18.23}{0.21} & \ms{56.22}{0.01} & \ms{56.53}{0.17} & \ms{21.18}{0.01} & \ms{54.36}{1.19} & \ms{54.11}{0.89} & \ms{20.03}{1.05} \\ 

\bottomrule
\end{tabular}
\end{adjustbox}
\label{tab:lp-trand-large-his}
\end{table*}

\begin{table*}[!h]
\caption{Temporal link prediction results on large-scale datasets under inductive settings with historical negative sampling. Results are averaged over three independent runs (in \%). }
\centering
\begin{adjustbox}{width=0.98\textwidth}
\begin{tabular}{cccccccccc}
\toprule
\multicolumn{1}{c}{\multirow{2}{*}{Methods}} & \multicolumn{3}{c}{Beeradvocate} & \multicolumn{3}{c}{Amazon-Kindle} & \multicolumn{3}{c}{Ratebeer} \\
\multicolumn{1}{c}{} & AP & AUROC & MRR & AP & AUROC & MRR & AP & AUROC & MRR \\
\midrule

JODIE & \ms{53.42}{0.25} & \ms{53.05}{0.41} & \ms{19.51}{0.14} & \ms{78.75}{0.39} & \ms{75.38}{0.08} & \ms{47.83}{0.95} & \ms{53.14}{0.30} & \ms{52.69}{0.53} & \ms{19.28}{0.44} \\ 
DyRep & \ms{50.82}{0.86} & \ms{49.19}{0.43} & \ms{17.78}{0.78} & \ms{73.53}{1.42} & \ms{68.86}{1.20} & \ms{41.20}{2.08} & \ms{54.15}{0.03} & \ms{53.32}{0.88} & \ms{20.03}{0.20} \\ 
TGAT & \ms{54.06}{0.47} & \ms{52.84}{0.14} & \ms{19.76}{0.37} & \ms{56.59}{0.09} & \ms{56.15}{0.15} & \ms{21.88}{0.05} & \ms{52.97}{0.79} & \ms{52.02}{0.25} & \ms{19.04}{0.68} \\ 
TGN & \ms{49.39}{0.31} & \ms{48.17}{0.39} & \ms{16.48}{0.23} & \ms{63.52}{0.44} & \ms{58.78}{0.36} & \ms{29.33}{0.76} & \ms{50.70}{0.90} & \ms{48.66}{0.41} & \ms{17.66}{0.79} \\ 
CAWN & \ms{52.81}{0.17} & \ms{53.49}{0.22} & \ms{18.60}{0.12} & \ms{56.49}{0.47} & \ms{55.43}{0.34} & \ms{21.89}{0.41} & \ms{54.89}{0.16} & \ms{56.34}{0.14} & \ms{19.94}{0.12} \\ 
TCL & \ms{54.23}{0.33} & \ms{52.83}{0.14} & \ms{20.06}{0.18} & \ms{55.04}{0.35} & \ms{54.30}{0.27} & \ms{20.66}{0.28} & \ms{56.75}{0.76} & \ms{55.37}{0.69} & \ms{22.38}{0.55} \\ 
GraphMixer & \ms{55.63}{1.05} & \ms{56.63}{0.78} & \ms{20.74}{0.88} & \ms{78.70}{0.73} & \ms{76.63}{0.39} & \ms{46.89}{1.27} & \ms{58.13}{0.03} & \ms{59.04}{0.21} & \ms{22.71}{0.12} \\ 
DyGFormer & \ms{55.95}{0.21} & \ms{54.07}{0.19} & \ms{21.27}{0.20} & \ms{56.36}{0.52} & \ms{55.74}{0.50} & \ms{21.55}{0.37} & \ms{57.23}{0.11} & \ms{57.84}{0.20} & \ms{21.90}{0.20} \\ 
FreeDyG & \ms{54.44}{0.68} & \ms{56.03}{0.00} & \ms{19.73}{0.62} & \ms{81.28}{0.69} & \ms{78.73}{0.51} & \ms{51.06}{1.32} & \ms{60.23}{0.56} & \ms{62.79}{0.98} & \ms{23.98}{0.34} \\ 
LKD4DyTAG & \ms{51.19}{0.05} & \ms{50.61}{0.13} & \ms{17.70}{0.01} & \ms{56.21}{0.10} & \ms{55.15}{0.07} & \ms{21.63}{0.09} & \ms{51.12}{0.25} & \ms{50.64}{0.35} & \ms{17.64}{0.17} \\ 
CROSS & \ms{54.58}{1.13} & \ms{52.96}{0.70} & \ms{20.32}{0.81} & \ms{59.80}{0.01} & \ms{59.52}{0.16} & \ms{24.25}{0.07} & \ms{58.00}{1.19} & \ms{58.44}{0.76} & \ms{22.67}{1.14} \\ 

\bottomrule
\end{tabular}
\end{adjustbox}
\label{tab:lp-ind-large-his}
\end{table*}

\begin{table*}[h]
\caption{Temporal link prediction results on small-scale datasets under transductive settings with inductive negative sampling. Results are averaged over three independent runs (in \%). }
\centering
\begin{adjustbox}{width=0.98\textwidth}
\begin{tabular}{cccccccccc}
\toprule
\multicolumn{1}{c}{\multirow{2}{*}{Methods}} & \multicolumn{3}{c}{FOOD} & \multicolumn{3}{c}{IMDB} & \multicolumn{3}{c}{Librarything} \\
\multicolumn{1}{c}{} & AP & AUROC & MRR & AP & AUROC & MRR & AP & AUROC & MRR \\
\midrule

JODIE & \ms{52.27}{1.48} & \ms{51.54}{0.87} & \ms{19.00}{1.18} & \ms{54.04}{9.34} & \ms{57.97}{13.71} & \ms{19.38}{7.05} & \ms{57.33}{1.47} & \ms{57.08}{1.90} & \ms{23.53}{1.14} \\ 
DyRep & \ms{55.10}{0.51} & \ms{52.88}{0.78} & \ms{21.37}{0.36} & \ms{57.29}{3.41} & \ms{50.87}{3.44} & \ms{26.33}{3.64} & \ms{63.21}{1.50} & \ms{63.42}{1.17} & \ms{28.00}{1.70} \\ 
TGAT & \ms{51.24}{0.20} & \ms{50.62}{0.21} & \ms{17.71}{0.21} & \ms{44.95}{0.38} & \ms{43.83}{1.01} & \ms{13.27}{0.12} & \ms{43.18}{0.24} & \ms{41.18}{0.70} & \ms{12.04}{0.06} \\ 
TGN & \ms{55.34}{0.24} & \ms{56.03}{0.49} & \ms{20.67}{0.10} & \ms{35.32}{0.12} & \ms{20.45}{0.41} & \ms{6.44}{0.14} & \ms{59.17}{0.46} & \ms{60.54}{0.29} & \ms{23.69}{0.45} \\ 
CAWN & \ms{53.06}{0.07} & \ms{53.83}{0.10} & \ms{18.77}{0.04} & \ms{46.32}{0.41} & \ms{44.02}{0.38} & \ms{14.62}{0.29} & \ms{47.85}{0.06} & \ms{48.47}{0.06} & \ms{14.92}{0.11} \\ 
TCL & \ms{50.53}{0.01} & \ms{50.22}{0.21} & \ms{17.13}{0.12} & \ms{46.19}{1.44} & \ms{44.56}{1.49} & \ms{14.55}{1.24} & \ms{43.74}{0.11} & \ms{42.97}{0.46} & \ms{11.76}{0.09} \\ 
GraphMixer & \ms{53.81}{0.68} & \ms{50.34}{1.35} & \ms{20.70}{0.30} & \ms{42.63}{0.27} & \ms{40.60}{0.85} & \ms{11.52}{0.05} & \ms{45.64}{0.55} & \ms{41.81}{0.15} & \ms{14.65}{0.46} \\ 
DyGFormer & \ms{49.32}{1.83} & \ms{47.18}{3.92} & \ms{16.67}{0.79} & \ms{46.97}{0.17} & \ms{44.98}{0.25} & \ms{15.29}{0.14} & \ms{49.46}{0.01} & \ms{51.30}{1.58} & \ms{15.74}{0.38} \\ 
FreeDyG & \ms{53.45}{0.38} & \ms{49.25}{0.98} & \ms{20.80}{0.18} & \ms{44.83}{0.08} & \ms{43.89}{0.16} & \ms{13.00}{0.01} & \ms{45.92}{0.42} & \ms{42.00}{0.04} & \ms{15.11}{0.40} \\ 
LKD4DyTAG & \ms{52.74}{0.10} & \ms{52.29}{0.13} & \ms{18.98}{0.09} & \ms{46.33}{0.89} & \ms{43.57}{0.76} & \ms{14.75}{0.78} & \ms{44.29}{0.06} & \ms{43.15}{0.08} & \ms{12.64}{0.02} \\ 
CROSS & \ms{50.92}{0.10} & \ms{50.07}{0.03} & \ms{17.54}{0.10} & \ms{48.93}{0.08} & \ms{47.19}{0.07} & \ms{16.50}{0.05} & \ms{46.54}{0.04} & \ms{47.15}{0.11} & \ms{13.82}{0.07} \\   

\bottomrule
\end{tabular}
\end{adjustbox}
\label{tab:lp-trand-small-ind}
\end{table*}

\begin{table*}[h]
\caption{Temporal link prediction results on small-scale datasets under inductive settings with inductive negative sampling. Results are averaged over three independent runs (in \%). }
\centering
\begin{adjustbox}{width=0.98\textwidth}
\begin{tabular}{cccccccccc}
\toprule
\multicolumn{1}{c}{\multirow{2}{*}{Methods}} & \multicolumn{3}{c}{FOOD} & \multicolumn{3}{c}{IMDB} & \multicolumn{3}{c}{Librarything} \\
\multicolumn{1}{c}{} & AP & AUROC & MRR & AP & AUROC & MRR & AP & AUROC & MRR \\
\midrule

JODIE & \ms{50.08}{0.39} & \ms{46.65}{0.10} & \ms{17.65}{0.47} & \ms{74.19}{0.85} & \ms{71.07}{1.83} & \ms{41.40}{0.35} & \ms{52.27}{0.06} & \ms{47.60}{0.69} & \ms{19.80}{0.09} \\ 
DyRep & \ms{48.32}{0.29} & \ms{45.69}{0.06} & \ms{16.11}{0.29} & \ms{69.73}{0.52} & \ms{65.67}{1.27} & \ms{36.57}{0.31} & \ms{50.86}{0.17} & \ms{46.70}{0.74} & \ms{18.40}{0.04} \\ 
TGAT & \ms{51.01}{0.25} & \ms{50.25}{0.03} & \ms{17.66}{0.30} & \ms{64.16}{0.67} & \ms{60.85}{0.71} & \ms{30.34}{0.56} & \ms{47.20}{0.22} & \ms{45.20}{0.09} & \ms{15.05}{0.15} \\ 
TGN & \ms{49.07}{0.34} & \ms{46.38}{0.32} & \ms{16.72}{0.23} & \ms{69.85}{2.88} & \ms{64.18}{1.71} & \ms{37.97}{4.38} & \ms{46.31}{0.19} & \ms{43.16}{0.34} & \ms{14.64}{0.16} \\ 
CAWN & \ms{54.97}{0.22} & \ms{53.61}{0.08} & \ms{20.31}{0.22} & \ms{55.21}{1.39} & \ms{54.88}{1.44} & \ms{21.38}{1.18} & \ms{50.89}{1.13} & \ms{50.06}{1.36} & \ms{17.55}{0.81} \\ 
TCL & \ms{53.10}{0.17} & \ms{49.76}{0.48} & \ms{19.31}{0.30} & \ms{57.06}{4.53} & \ms{54.56}{2.23} & \ms{23.76}{4.87} & \ms{47.35}{0.47} & \ms{44.99}{0.48} & \ms{15.14}{0.41} \\ 
GraphMixer & \ms{56.46}{1.19} & \ms{53.37}{1.60} & \ms{22.80}{0.94} & \ms{75.58}{0.02} & \ms{71.56}{0.10} & \ms{44.36}{0.21} & \ms{61.46}{2.92} & \ms{59.97}{2.19} & \ms{26.70}{2.84} \\ 
DyGFormer & \ms{48.37}{1.76} & \ms{46.33}{3.88} & \ms{15.92}{0.80} & \ms{53.50}{0.41} & \ms{53.75}{0.04} & \ms{19.53}{0.49} & \ms{53.57}{0.32} & \ms{56.43}{0.14} & \ms{18.51}{0.29} \\ 
FreeDyG & \ms{54.50}{0.49} & \ms{50.54}{1.00} & \ms{21.36}{0.24} & \ms{76.96}{0.31} & \ms{73.35}{0.45} & \ms{46.00}{0.55} & \ms{58.12}{1.22} & \ms{56.72}{1.11} & \ms{23.62}{1.13} \\ 
LKD4DyTAG & \ms{50.34}{0.08} & \ms{49.50}{0.13} & \ms{17.12}{0.01} & \ms{51.36}{0.19} & \ms{51.15}{0.01} & \ms{17.98}{0.19} & \ms{48.17}{0.04} & \ms{46.87}{0.00} & \ms{15.57}{0.01} \\ 
CROSS & \ms{49.78}{0.42} & \ms{48.83}{0.34} & \ms{16.78}{0.33} & \ms{55.17}{1.15} & \ms{56.38}{0.67} & \ms{20.51}{1.17} & \ms{50.15}{0.24} & \ms{50.27}{0.43} & \ms{ 16.61}{0.14} \\ 

\bottomrule
\end{tabular}
\end{adjustbox}
\label{tab:lp-ind-small-ind}
\end{table*}

\subsection{More Experiment Results and Analysis}
\label{appx:more-results}


\textbf{Detailed Main Results.}  Regarding the TLP task, \cref{tab:flp-detail-1,tab:flp-detail-2} supplement \cref{tab:main-lp} by providing results for an extended set of evaluation metrics. 
 
\textbf{TLP Inductive Results.} In line with the evaluation protocol of DyGLib~\cite{yu2023towards}, we also assess the performance of all methods under inductive scenarios in addition to transductive settings for TLP task. The transductive setting focuses on predicting future links among nodes encountered during training, whereas the inductive setting requires predicting future links for previously unseen nodes. The detailed inductive results for temporal link prediction task  are presented in~\cref{tab:lp-ind-small,tab:lp-ind-large}.

\textbf{Additional Negative Sampling Results.}  
Following DyGLib~\cite{yu2023towards}, we additionally evaluate all methods on the TLP task under both historical and inductive negative sampling strategies. The results are reported in \cref{tab:lp-trand-small-his,tab:lp-trand-large-his,tab:lp-trand-small-ind,tab:lp-trand-large-ind,tab:lp-ind-small-his,tab:lp-ind-small-ind,tab:lp-ind-large-his,tab:lp-ind-large-ind}.

\begin{figure}[h]
  \centering
  \includegraphics[width=\linewidth]{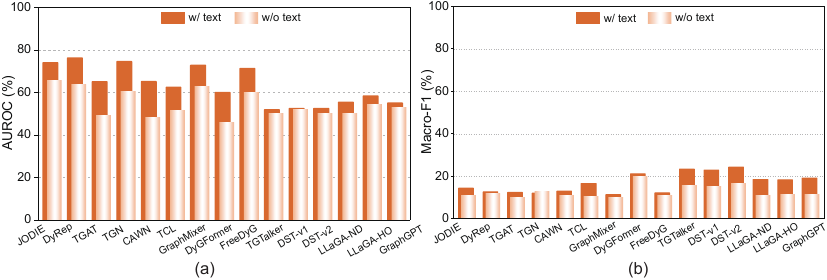}
  \caption{Effect of Text Attributes on TLP. Impact of textual attributes on model performance across different methods for the TLP task.}
  \label{fig:flp-text}
\end{figure}

\textbf{Text Ablation Study Results.}  
\cref{fig:flp-text} presents detailed performance comparisons on both TLP and TNC before and after removing textual attributes. The results show that textual information substantially affects TGNN performance on TLP but has relatively limited impact on TNC, whereas for LLM-Predictors the opposite trend holds: text contributes significantly to TNC while providing smaller gains for TLP.

\textbf{Comparison of LP and E2E for TGNNs on TNC.}  
As shown in \cref{tab:lp-e2e}, end-to-end fine-tuning does not yield significant improvements over linear probing on the TNC task for TGNNs. This suggests that gradients from semantic supervision are insufficient to overcome the strong structural bias learned during TLP pretraining. Consequently, the representations remain anchored in topological dynamics and fail to adapt effectively to semantic objectives.


\begin{figure}[!h]
  \centering
  \includegraphics[width=\linewidth]{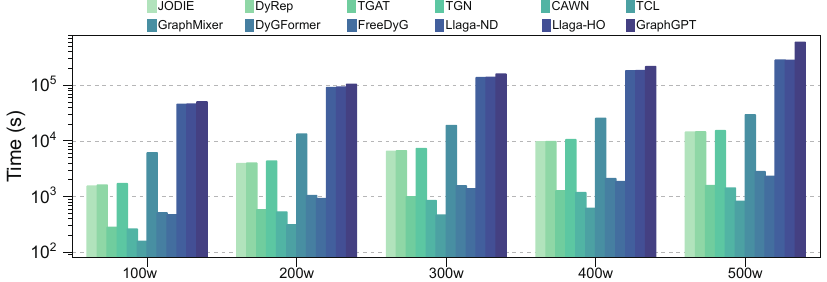}
  \caption{Scalability results across all methods.}
  \label{fig:scalability}
\end{figure}

\textbf{Scalability Results.}  
\cref{fig:scalability} reports scalability evaluations of all methods. Specifically, on the largest dataset, Amazon-Kindle, we subsample 1M, 2M, 3M, 4M, and 5M edges and measure the runtime of each method. Results show that all methods exhibit approximately linear growth in computation time as data size increases, demonstrating good scalability. LLM-Predictor methods incur higher time cost due to LLM inference, but still maintain favorable scaling behavior.

\begin{table*}[!h]
\caption{Temporal link prediction results on large-scale datasets under transductive settings with inductive negative sampling. Results are averaged over three independent runs (in \%). }
\centering
\begin{adjustbox}{width=0.98\textwidth}
\begin{tabular}{cccccccccc}
\toprule
\multicolumn{1}{c}{\multirow{2}{*}{Methods}} & \multicolumn{3}{c}{Beeradvocate} & \multicolumn{3}{c}{Amazon-Kindle} & \multicolumn{3}{c}{Ratebeer} \\
\multicolumn{1}{c}{} & AP & AUROC & MRR & AP & AUROC & MRR & AP & AUROC & MRR \\
\midrule

JODIE & \ms{57.75}{2.59} & \ms{63.12}{0.18} & \ms{23.14}{3.13} & \ms{58.69}{2.39} & \ms{55.79}{2.66} & \ms{25.10}{2.06} & \ms{66.28}{0.28} & \ms{70.43}{0.20} & \ms{30.59}{0.44} \\ 
DyRep & \ms{71.73}{2.56} & \ms{74.04}{2.53} & \ms{38.10}{2.50} & \ms{62.65}{1.46} & \ms{61.27}{1.96} & \ms{27.58}{1.08} & \ms{72.86}{1.78} & \ms{75.20}{2.50} & \ms{38.07}{1.33} \\ 
TGAT & \ms{50.57}{0.25} & \ms{50.73}{0.16} & \ms{17.17}{0.28} & \ms{48.21}{0.78} & \ms{47.43}{1.01} & \ms{16.15}{0.66} & \ms{49.80}{0.22} & \ms{49.28}{0.25} & \ms{16.90}{0.23} \\ 
TGN & \ms{54.88}{0.29} & \ms{60.31}{1.15} & \ms{20.32}{0.72} & \ms{35.99}{0.65} & \ms{23.46}{2.47} & \ms{7.83}{0.52} & \ms{64.44}{3.75} & \ms{71.05}{2.62} & \ms{27.93}{3.69} \\ 
CAWN & \ms{51.43}{0.06} & \ms{52.82}{0.10} & \ms{17.77}{0.04} & \ms{49.99}{1.19} & \ms{49.20}{1.82} & \ms{17.48}{0.79} & \ms{52.54}{0.12} & \ms{55.03}{0.13} & \ms{18.23}{0.18} \\ 
TCL & \ms{50.23}{0.39} & \ms{50.38}{0.23} & \ms{16.77}{0.33} & \ms{49.56}{0.54} & \ms{47.92}{0.81} & \ms{17.44}{0.40} & \ms{50.93}{0.85} & \ms{50.36}{0.63} & \ms{17.65}{0.89} \\ 
GraphMixer & \ms{51.71}{0.75} & \ms{52.29}{0.73} & \ms{19.50}{0.58} & \ms{50.51}{0.13} & \ms{49.41}{0.07} & \ms{18.45}{0.14} & \ms{52.58}{0.20} & \ms{52.42}{0.44} & \ms{19.84}{0.39} \\ 
DyGFormer & \ms{50.30}{0.33} & \ms{50.62}{0.15} & \ms{16.80}{0.27} & \ms{50.89}{0.94} & \ms{50.79}{0.94} & \ms{17.57}{0.64} & \ms{52.44}{0.20} & \ms{53.81}{0.34} & \ms{18.13}{0.13} \\ 
FreeDyG & \ms{51.41}{0.41} & \ms{52.26}{0.24} & \ms{19.03}{0.61} & \ms{51.03}{0.02} & \ms{49.42}{0.38} & \ms{18.19}{0.03} & \ms{54.36}{1.00} & \ms{56.53}{1.66} & \ms{20.11}{0.69} \\ 
LKD4DyTAG & \ms{50.53}{0.07} & \ms{50.64}{0.09} & \ms{17.93}{0.12} & \ms{51.88}{0.71} & \ms{50.77}{1.01} & \ms{19.03}{0.48} & \ms{49.62}{0.08} & \ms{49.11}{0.19} & \ms{16.92}{0.05} \\ 
CROSS & \ms{50.13}{0.39} & \ms{50.02}{0.57} & \ms{16.81}{0.20} & \ms{55.64}{0.05} & \ms{56.70}{0.47} & \ms{21.64}{0.10} & \ms{53.31}{0.73} & \ms{54.90}{0.66} & \ms{18.83}{0.64} \\

\bottomrule
\end{tabular}
\end{adjustbox}
\label{tab:lp-trand-large-ind}
\end{table*}

\begin{table*}[!h]
\caption{Temporal link prediction results on large-scale datasets under inductive settings with inductive negative sampling. Results are averaged over three independent runs (in \%). }
\centering
\begin{adjustbox}{width=0.98\textwidth}
\begin{tabular}{cccccccccc}
\toprule
\multicolumn{1}{c}{\multirow{2}{*}{Methods}} & \multicolumn{3}{c}{Beeradvocate} & \multicolumn{3}{c}{Amazon-Kindle} & \multicolumn{3}{c}{Ratebeer} \\
\multicolumn{1}{c}{} & AP & AUROC & MRR & AP & AUROC & MRR & AP & AUROC & MRR \\
\midrule

JODIE & \ms{53.42}{0.25} & \ms{53.05}{0.41} & \ms{19.51}{0.14} & \ms{78.75}{0.39} & \ms{75.38}{0.08} & \ms{47.83}{0.95} & \ms{ 53.14}{0.30} & \ms{52.69}{0.53} & \ms{ 19.28}{0.44} \\ 
DyRep & \ms{50.82}{0.86} & \ms{49.19}{0.43} & \ms{17.78}{0.78} & \ms{73.53}{1.42} & \ms{68.86}{1.20} & \ms{ 41.20}{2.08} & \ms{54.15}{0.03} & \ms{53.32}{0.88} & \ms{20.03}{0.20} \\ 
TGAT & \ms{54.06}{0.47} & \ms{52.84}{0.14} & \ms{19.76}{0.37} & \ms{56.59}{0.09} & \ms{56.15}{0.15} & \ms{21.88}{0.05} & \ms{52.97}{0.79} & \ms{ 52.02}{0.25} & \ms{ 19.04}{0.68} \\ 
TGN & \ms{49.39}{0.31} & \ms{48.17}{0.39} & \ms{ 16.48}{0.23} & \ms{63.52}{0.44} & \ms{58.78}{0.36} & \ms{29.33}{0.76} & \ms{ 50.70}{0.90} & \ms{ 48.66}{0.41} & \ms{ 17.66}{0.79} \\ 
CAWN & \ms{ 52.81}{0.17} & \ms{ 53.49}{0.22} & \ms{ 18.60}{0.12} & \ms{ 56.49}{0.47} & \ms{55.43}{0.34} & \ms{21.89}{0.41} & \ms{54.89}{0.16} & \ms{56.34}{0.14} & \ms{19.94}{0.12} \\ 
TCL & \ms{54.23}{0.33} & \ms{52.83}{0.14} & \ms{20.06}{0.18} & \ms{55.04}{0.35} & \ms{54.30}{0.27} & \ms{20.66}{0.28} & \ms{56.75}{0.76} & \ms{55.37}{0.69} & \ms{22.38}{0.55} \\ 
GraphMixer & \ms{55.63}{1.05} & \ms{56.63}{0.78} & \ms{20.74}{0.88} & \ms{78.70}{0.73} & \ms{76.63}{0.39} & \ms{46.89}{1.27} & \ms{58.13}{0.03} & \ms{59.04}{0.21} & \ms{22.71}{0.12} \\ 
DyGFormer & \ms{55.95}{0.21} & \ms{54.07}{0.19} & \ms{21.27}{0.20} & \ms{56.36}{0.52} & \ms{55.74}{0.50} & \ms{21.55}{0.37} & \ms{57.23}{0.11} & \ms{57.84}{0.20} & \ms{21.90}{0.20} \\ 
FreeDyG & \ms{54.44}{0.68} & \ms{56.03}{0.00} & \ms{19.73}{0.62} & \ms{81.28}{0.69} & \ms{78.73}{0.51} & \ms{51.06}{1.32} & \ms{60.23}{0.56} & \ms{62.79}{0.98} & \ms{23.98}{0.34} \\ 
LKD4DyTAG & \ms{51.19}{0.05} & \ms{50.61}{0.13} & \ms{17.70}{0.01} & \ms{56.21}{0.10} & \ms{55.15}{0.07} & \ms{21.63}{0.09} & \ms{51.12}{0.25} & \ms{50.64}{0.35} & \ms{17.64}{0.17} \\ 
CROSS & \ms{54.58}{1.13} & \ms{52.96}{0.70} & \ms{20.32}{0.81} & \ms{59.80}{0.01} & \ms{59.52}{0.16} & \ms{24.25}{0.07} & \ms{58.00}{1.19} & \ms{58.44}{0.76} & \ms{22.67}{1.14} \\  

\bottomrule
\end{tabular}
\end{adjustbox}
\label{tab:lp-ind-large-ind}
\end{table*}

\begin{table*}[h]
\caption{Performance comparison of TGNNs on the TNC task under two training strategies: linear probing (LP) and end-to-end fine-tuning (E2E). Linear probing trains only the classifier head on a TLP-pretrained backbone, while end-to-end fine-tuning jointly optimizes both the backbone and the classifier. Results are reported in terms of Macro-F1.}
\centering
\begin{adjustbox}{width=0.72\textwidth}
\begin{tabular}{ccccccccccc}
\toprule
\multicolumn{1}{c}{\multirow{2}{*}{Methods}} & \multicolumn{2}{c}{FOOD} & \multicolumn{2}{c}{IMDB} & \multicolumn{2}{c}{Beeradvocate} & \multicolumn{2}{c}{Amazon-Kindle} & \multicolumn{2}{c}{Ratebeer}\\
\multicolumn{1}{c}{} & LP & E2E & LP & E2E & LP & E2E & LP & E2E & LP & E2E \\
\midrule
JODIE & 25.95 & 25.94 & 12.65 & 12.65 & 7.00  & 6.95  & 17.60 & 13.18 & 7.86  & 8.03  \\ 
DyRep & 21.22 & 21.40 & 13.74 & 13.77 & 6.50  & 6.49  & 13.36 & 14.02 & 7.38  & 8.62  \\ 
TGAT & 22.73 & 22.73 & 8.53  & 8.53  & 6.18  & 6.14  & 14.91 & 15.34 & 8.67  & 8.66  \\ 
TGN & 21.98 & 21.80 & 8.51  & 8.51  & 8.35  & 7.95  & 13.22 & OOM & 7.34  & 8.22  \\ 
CAWN & 20.28 & 19.99 & 11.48 & 11.49 & 8.77  & 8.66  & 14.96 & 15.96 & 8.79  & 9.04  \\ 
TCL & 19.99 & 19.99 & 14.56 & 13.60 & 9.26  & 8.17  & 24.25 & 27.06 & 14.34 & 14.45 \\ 
GraphMixer & 19.99 & 19.99 & 9.20  & 9.29  & 6.14  & 6.14  & 13.35 & 14.62 & 7.13  & 7.27  \\ 
DyGFormer & 36.58 & 26.58 & 20.05 & 12.85 & 12.91 & 10.16 & 20.19 & 19.16 & 15.52 & 13.79 \\ 
FreeDyG & 20.46 & 20.88 & 10.20 & 10.02 & 7.75  & 7.24  & 12.80 & 13.77 & 8.54  & 8.28 \\ 

\bottomrule
\end{tabular}
\end{adjustbox}
\label{tab:lp-e2e}
\end{table*}

\end{document}